\documentclass[review]{elsarticle}

\pdfoutput=1 
\usepackage[colorlinks=true, linkcolor=black, citecolor=black, urlcolor=black]{hyperref}
\usepackage{subfigure}
\usepackage[usenames]{color}
\usepackage[numbers]{natbib}
\usepackage{verbatim}
\usepackage{lineno}

\usepackage[figuresright]{rotating}
\usepackage[table]{xcolor}
\usepackage{amssymb}
\usepackage{makecell}
\usepackage{multirow}
\usepackage{rotating}
\usepackage{array}
\usepackage{times}
\usepackage{psfrag}
\usepackage{subfigure}
\usepackage{rotating}
\usepackage{amsmath}
\usepackage{booktabs}
\usepackage{lscape}
\usepackage{verbatim}
\usepackage{geometry}
\usepackage[misc]{ifsym}
\usepackage{capt-of}
\usepackage{pdflscape}
\usepackage{afterpage}
\usepackage{floatrow}
\usepackage[labelsep=space]{caption}
\usepackage{pdflscape}

\journal{Journal of \LaTeX\ Templates}

\begin{document}

\begin{frontmatter}

\title{New Ideas and Trends in Deep Multimodal \\ Content Understanding: A Review}

\author[label1]{Wei Chen}
\author[label2]{Weiping Wang}
\author[label2,label3]{Li Liu}
\author{Michael S. Lew\corref{cor1}\fnref{label1}}
\cortext[cor1]{Corresponding author}
\ead{m.s.k.lew@liacs.leidenuniv.nl}
\address[label1]{LIACS, Leiden University, Leiden, 2333 CA, The Netherlands}
\address[label2]{College of Systems Engineering, NUDT, Changsha, 410073, China}
\address[label3]{Center for Machine Vision and Signal Analysis, University of Oulu, Finland}

\begin{abstract}

The focus of this survey is on the analysis of two modalities of multimodal deep learning: image and text. Unlike classic reviews of deep learning where monomodal image classifiers such as VGG, ResNet and Inception module are central topics, this paper will examine recent multimodal deep models and structures, including auto-encoders, generative adversarial nets and their variants. These models go beyond the simple image classifiers in which they can do uni-directional (\emph{e.g.} image captioning, image generation) and bi-directional (\emph{e.g.} cross-modal retrieval, visual question answering) multimodal tasks. Besides, we analyze two aspects of the challenge in terms of better content understanding in deep multimodal applications. We then introduce current ideas and trends in deep multimodal feature learning, such as feature embedding approaches and objective function design, which are crucial in overcoming the aforementioned challenges. Finally, we include several promising directions for future research. 

\end{abstract}


\end{frontmatter}


\section*{1 Introduction}
\label{section1}

Semantic information that helps us illustrate the world usually comes from different sensory modalities in which the event is processed or is experienced (\emph{i.e.} auditory, tactile, or visual). Thus, the same concept or scene can be presented in different ways. If we consider a scene where ``\emph{a large yellow dog leaps into the air to catch a frisbee}'', then one could select audio or video or an image, which also indicates the multimodal aspect of the problem. To perform multimodal tasks well, first, it is necessary to understand the content of multiple modalities. Multimodal content understanding aims at recognizing and localizing objects, determining the attributes of objects, characterizing the relationships between objects, and finally, describing the common semantic content among different modalities. In the information era, rapidly developing technology makes it more convenient than ever to access a sea of multimedia data such as text, image, video, and audio. As a result, exploring semantic correlation to understand content for diverse multimedia data has been attracting much attention as a long-standing research field in the computer vision community.

\textcolor{black}{Recently, the topics range from speech-video to image-text applications. Considering the wide array of topics, we restrict the scope of this survey to image and text data specifically in the multimodal research community, including tasks at the intersection of image and text (also called cross-modal). According to the available modality during testing stage, multimodal applications include bi-directional tasks (\emph{e.g.} image-sentence search \cite{park2018retrieval}\cite{mandal2019generalized}, visual question answering (VQA) \cite{liang2019focal}\cite{wang2018fvqa}) and uni-directional tasks (\emph{e.g.} image captioning \cite{chen2018show}\cite{wu2018image}, image generation \cite{cha2019adversarial}\cite{reed2016generative}), both of them will be introduced in the following sections.}

\begin{figure*}[t]
\centering
\includegraphics[width=0.8\textwidth]{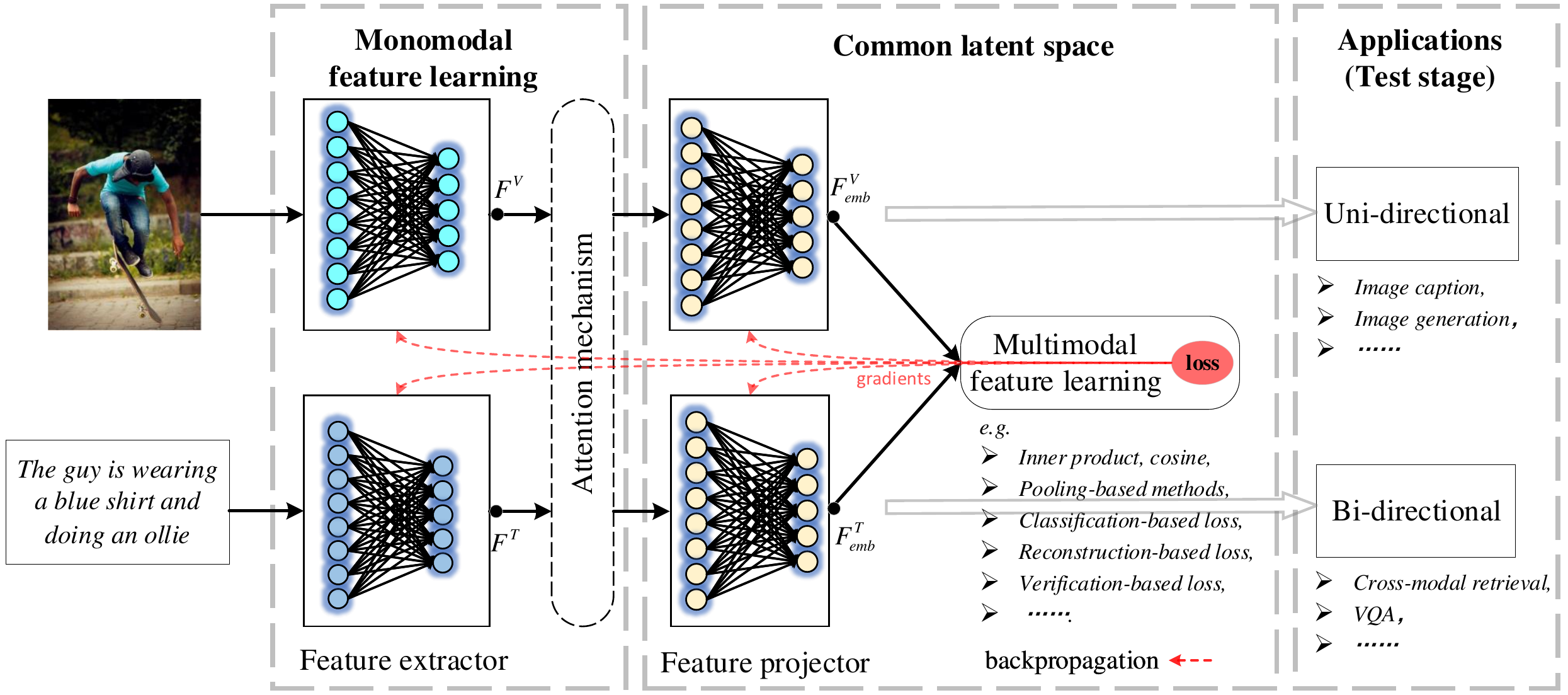}
\caption{ \textcolor{black}{A general flowchart of deep multimodal feature learning. Each modality starts with using an individual neural network to process the data (\emph{e.g.} CNN for images and RNN for text), which implements monomodal feature learning. The attention module is an optional module for aligning two monomodal features. The extracted features $ F^{V} $ and $ F^{T}$ are not directly comparable and are distributed inconsistently due to the process of individual domain-specific neural networks. To understand the multimodal content, these monomodal features are embedded into a common latent space with the help of mapping functions (\emph{e.g.} MLP). According to the taxonomy \cite{baltruvsaitis2019multimodal}, feature embedding in the common space can be categorized into a joint and a coordinated representation. Afterwards, the optimized multimodal features $ F_{emb}^{T} $ and $ F_{emb}^{V} $ are comparable for specific applications. Common latent feature learning is a crucial module in deep multimodal learning, which aims at learning semantically discriminative features. The current ideas and trends for improving performance are the central topic of this paper. }\label{The_pipeline_of_deep_multimodal_research}}
\end{figure*}

\textcolor{black}{With the powerful capabilities of deep neural networks, data from visual and textual modality can be represented as individual features using domain-specific neural networks. Complementary information from these unimodal features is appealing for multimodal content understanding. For example, the individual features can be further projected into a common space by using another neural network for a prediction task. For clarity, we illustrate the flowchart of neural networks for multimodal research in Figure \ref{The_pipeline_of_deep_multimodal_research}. On the one hand, the neural networks are comprised by successive linear layers and non-linear activation functions, the image or text data is represented in a high abstraction way, which leads to the ``semantic gap'' \cite{li2016socializing}. On the other hand, different modalities are characterized by different statistical properties. Image is 3-channel RGB array while text is often symbolic. When represented by different neural networks, their features have unique distributions and differences, which leads to the ``heterogeneity gap'' \cite{wang2016comprehensive}.
That is to say, to understand multimodal content, deep neural networks should be able to reduce the difference between high-level semantic concepts and low-level features in intra-modality representations, as well as construct a common latent space to capture semantic correlations in inter-modality representations. }

Much effort has gone into mitigating these two challenges to improve content understanding. Some works involve deep multimodal structures such as cycle-consistent reconstruction \cite{Gorti2018Text}\cite{wu2019cycle}\cite{Liu2018SemanticIS}, while others focus on feature extraction nets such as graph convolutional networks \cite{Yu2018TextualRM}\cite{Yao2018ExploringVR}\cite{Yang2018AutoEncodingSG}. In some algorithms, reinforcement learning is combined with deep multimodal feature learning \cite{rennie2017self}\cite{liu2018context}\cite{liu2018show}. These recent ideas are the scope of this survey. \textcolor{black}{In a previous review \cite{baltruvsaitis2019multimodal}, the authors analyze intrinsic issues for multimodal research but mainly focus on machine learning.} Some recent advances in deep multimodal feature learning are introduced \cite{ramachandram2017deep}, but it mainly discusses feature fusion structures and regularization strategies. 

In this paper, we focus on two specific modalities, image and text, by examining recent related ideas. First, we focus on the structures of deep multimodal models, including auto-encoders and generative adversarial networks  \cite{goodfellow2014generative} and their variants. These models, which perform uni-directional or bi-directional tasks, go beyond simple image classifiers (\emph{e.g.} ResNet). Second, we analyze recent methods of multimodal feature extraction which aim at getting semantically related features to minimize the heterogeneity gap. Third, we focus on current popular algorithms for common latent feature learning, which are beneficial for network training to preserve semantic correlations between modalities. In conclusion, the newly applied ideas mitigate the ``heterogeneity gap'' and the ``semantic gap'' between visual and textual modalities.

The rest of this paper is organized as follows: Section 2 introduces image-text related applications, followed by corresponding challenges and intrinsic issues for these image-text applications in Section 3. Regarding these challenges, we analyze the current ideas and trends in deep multimodal learning in Section 4. Finally, we conclude with several promising directions in Section 5.

\vspace{-1.5em}
\section*{2 Multimodal Applications}
\label{Multimodal_Applications}

\vspace{-1em}

This section aims to summarize various multimodal applications where image and text data are involved. These applications have gained a lot of attention lately and show a natural division into uni-directional and bi-directional groups. The difference is that for uni-directional scenarios only one modality is available at the test stage, whereas in bi-directional scenarios, two modalities are required. 

\subsection*{2.1 Uni-directional Multimodal Applications}

An important concern in deep multimodal research is to map (translate) one modality to another. For example, given an entity in visual (or textual) space, the task is to generate a description of this entity in textual (or visual) space according to the content. For some tasks, these mapping processes are uni-directional, \emph{i.e.} either from image to text or from text to image.

(1) Image-to-text tasks

Image captioning is a task that generates a sentence description for an image and requires recognizing important objects and their attributes, then inferring their correlations within the image \citep{hossain2019comprehensive}. After capturing these correlations, the captioner yields a syntactically correct and semantically relevant sentence. To understand the visual content, images are fed into convolutional neural networks to learn hierarchical features, which constitutes the feature encoding process. The produced hierarchical features are transformed into sequential models (\emph{e.g.} RNN, LSTM) to generate the corresponding descriptions. Subsequently, the evaluation module produces description difference as the feedback signals to update the performance of each block. Deep neural networks are commonly used in image captioning. For other methods, including retrieval- and template-based methods, we recommend the existing surveys \cite{hossain2019comprehensive}\cite{He2017Deep}\cite{bai2018survey}. In the following sections, we will examine the methods widely used to improve image captioning performance, including evolutionary algorithm \cite{wang2020evolutionary}, generative adversarial networks \cite{goodfellow2014generative}\cite{dai2017towards}\cite{shetty2017speaking}, reinforcement learning \cite{rennie2017self}\cite{liu2018context}\cite{liu2018show}, memory networks \cite{weston2014memory}\cite{sukhbaatar2015end}\cite{park2018towards}, and attention mechanisms \cite{anderson2018bottom}\cite{Wang2019HierarchicalAttention}\cite{Song2019connectingLanguage}\cite{wang2020learning}.

\textcolor{black}{Image captioning is an open-ended research question.
It is still difficult to evaluate the performance of captioning, which should be diverse, creative, and human-like \cite{cui2018learning}. Currently, the metrics for evaluating the performance of image captioning include \textit{BLEU} (Bilingual Evaluation Understanding), \textit{ROUGE} (Recall-Oriented Understudy for Gisting Evaluation), \textit{METEOR} (Metric for Evaluation of Translation with Explicit ORdering), \textit{CIDEr} (Consensus-based Image Description Evaluation), and \textit{SPICE} (Semantic Propositional Image Captioning Evaluation). It is hard to access the generated captions by linguists. Thus it is necessary to further study an evaluation indicator which is more in line with human judgments and is flexible to new pathological cases \cite{cui2018learning}. Furthermore, image captioning systems might suffer from dataset bias issue. The trained captioner overfits to the common objects in seen context (\emph{e.g.} book and desk), but it would be challenging for the captioner to generalize to the same objects in unseen contexts (\emph{e.g.} book and tree).
}

\textcolor{black}{According to captioning principles, researchers focus on specific caption generation tasks, such as image tagging \cite{jin2016annotation}, visual region captioning \cite{li2017scene}, and object captioning \cite{anderson2018partially}. Analogously, these tasks are also highly dependent on the regional image patch and sentences/phrases organization. The specific correlations between the features of objects (or regions) in one image and the word-level (or phrase-level) embeddings are explored instead of global dependence of the holistic visual and textual features.}


(2) Text-to-image tasks

Compared to generating a sentence for a given image, generating a realistic and plausible image from a sentence is even more challenging. Namely, it is difficult to capture semantic cues from a highly abstract text, especially when the text is used to describe complex scenarios as found in the MS-COCO dataset \cite{el2019gilt}\cite{lin2014microsoft}. Text-to-image generation is such a kind of task which maps from textual modality to visual modality.

Text-to-image generation requires synthesized images to be photo-realistic and semantically consistent (\emph{i.e.} preserving specific object sketches and semantic textures described in text data). Generally, this requirement is closely related to the following two aspects: the heterogeneity gap \cite{wang2016comprehensive} and the semantic gap \cite{li2016socializing}\cite{yuan2018text}. The first addresses the gap between the high-level concepts of text descriptions and the pixel-level values of an image, while the second exists between synthetic images and real images. 

The above issues in text-to-image application are exactly what generative models attempt to address, through methods such as Variational Auto-Encoders (VAE) \cite{mansimov2015generating}, auto-regressive models \cite{reed2017parallel} and Generative Adversarial Networks (GANs) \cite{reed2016generative}\cite{goodfellow2014generative}. Recently, various new ideas and network architectures have been proposed to improve image generation performance. One example is to generate a semantic layout as intermediate information from text data to bridge the heterogeneity gap in image and text \cite{Johnson2018Image}\cite{Hong2018Inferring}\cite{Tan2018Text2Scene}. Some works focus on the network structure design for feature learning. For image synthesis, novel derivative architectures from GANs \cite{He2018An} have been explored in hierarchically nested adversarial networks \cite{Zhang2018Photographic}, perceptual pyramid adversarial networks \cite{Gao2019Perceptual}, iterative stacked networks \cite{Han2016StackGAN}\cite{HanZ2017StackGAN}, attentional generative networks \cite{Tao18attngan}\cite{articleReed}, cycle-consistent adversarial networks \cite{Gorti2018Text}\cite{Liu2018SemanticIS} and symmetrical distillation networks \cite{yuan2018text}.

\textcolor{black}{ Image generation is a promising multimodal application and has many applicable scenarios such as photo editing or multimedia data creation. Thereby, this task has attracted lots of attention. However, there are two main limitations to be explored further. Similar with image captioning, the first limit is regarding the evaluation metrics. Currently, Inception Score (IS) \cite{Han2016StackGAN}\cite{HanZ2017StackGAN}\cite{zhang2019self}, Fr\'{e}chet Inception Distance (FID) \cite{zhang2019self}, Multi-scale Structural Similarity Index Metrics (MS-SSIM) \cite{snell2017learning}\cite{odena2017conditional}, and Visual-semantic Similarity (VS) \cite{Zhang2018Photographic} are used to evaluate generation quality. These metrics pay attention to generated image resolution and image diversity. However, performance is still far from human perception. Another limitation is that, while generation models work well and achieve promising results on single category object datasets like Caltech-UCSD CUB \cite{reed2016learning} and Oxford-102 Flower \cite{reed2016learning}, existing methods are still far from promising on complex dataset like MS-COCO where one image contains more objects and is described by a complex sentence.}

To compensate for these limitations, word-level attention \cite{Tao18attngan}, hierarchical text-to-image mapping \cite{Hong2018Inferring} and memory networks \cite{zhang2018text} have been explored. In the future, one direction may be to make use of the Capsule idea proposed by Hinton \cite{Sabour2017DynamicRB} since capsules are designed to capture the concepts of objects \cite{He2018An}.

\vspace{-1em}
\subsection*{2.2 Bi-directional Multimodal Applications}

\vspace{-0.5em}

As for bi-directional applications, features from visual modality are translated to textual modality and vice versa.
Representative bi-directional applications are cross-modal retrieval and visual question answering (VQA) where image and text are projected into a common space to explore their semantic correlations.

(1) Cross-modal retrieval

\textcolor{black}{Single-modal and cross-modal retrieval have been researched for decades \cite{peng2018overview}. Different from single-modal retrieval, cross-modal retrieval returns the most relevant image (text) when given a query text (image). As for performance evaluation, there are two important aspects:  retrieval accuracy and retrieval efficiency. }
   
\textcolor{black}{For the first, it is desirable to explore semantic correlations across an image and text features. To meet this requirement, the aforementioned heterogeneity gap and the semantic gap are the challenges to deal with. Some novel techniques that have been proposed are as follows: attention mechanisms and memory networks are employed to align relevant features between image and text \cite{huang2019bi}\cite{wang2018joint}\cite{song2018deep}\cite{yu2020learning}; Bi-directional sequential models (\emph{e.g.} Bi-LSTM \cite{graves2005bidirectional}) are used to explore spatial-semantic correlations \cite{park2018retrieval}\cite{huang2019bi}; Graph-based embedding and graph regularization are utilized to keep semantic order in text feature extraction process \cite{zhang2019supervised}\cite{wu2018learning}; Information theory is applied to reduce the heterogeneity gap in cross-modal hasing \cite{chen2019domain}; Adversarial learning strategies and GANs are used to estimate common feature distributions in cross-modal retrieval \cite{peng2019cm}\cite{wang2017adversarial}\cite{wu2020augmented}; Metric learning strategies are explored, which consider inter-modality semantic similarity and intra-modality neighborhood constraints \cite{wang2019learning}\cite{zhang2018deep}\cite{zhan2018comprehensive}.}
   
\textcolor{black}{For the second, recent hashing methods have been explored \cite{mandal2019generalized}\cite{jin2019deep}\cite{zhang2018collaborative}\cite{deng2018triplet}\cite{zhang2018attention}\cite{jiang2017deep}\cite{cao2018cross}\cite{zhang2018sch}\cite{li2018self}\cite{wang2020batch}\cite{yao2020efficient} owing to the computation and storage advantages of binary code. Essentially, methods such as attention mechanisms \cite{zhang2018attention} and adversarial learning \cite{zhang2018sch}\cite{li2018self}\cite{song2018binary} are applied for learning compact hash codes with different lengths. However, the problems should be considered when one employs hashing methods for cross-modal retrieval are feature quantization and non-differential binary code optimization. Some methods, such as self-supervised learning \cite{li2018self} and continuation  \cite{song2018binary}, are explored to address these two issues. Recently, \textcolor{black}{Yao et al. \cite{yao2020efficient} introduce an efficient discrete optimal scheme for binary code learning in which a hash codes matrix is construct.} Focusing on the feature quantization, Wang et al. \cite{wang2020batch} introduce a hashing code learning algorithm in which the binary codes are generated without relaxation so that the large quantization and non-differential problems are avoided. \textcolor{black}{Analogously, a straightforward discrete hashing code optimization strategy is proposed, more importantly, in an unsupervised  way \cite{wang2019unsupervised}.} }

Although much attention has been paid to cross-modal retrieval, there still exists room for performance improvement (see Figure \ref{Flickr30k_MSCOCO}-\ref{Flickr25k_NUSWIDE_Hashing}). For example, to employ graph-based methods to construct semantic information within two modalities, more context information such as objects link relationships are adopted for more effective semantic graph construction \cite{peng2018overview}.

(2) Visual question answering 

\begin {figure}[!t]
\centering
 \subfigure[``Yes/no'' problem]
 { \label{State_of_the_art_Holidays}     
   \includegraphics[width=0.486\columnwidth]{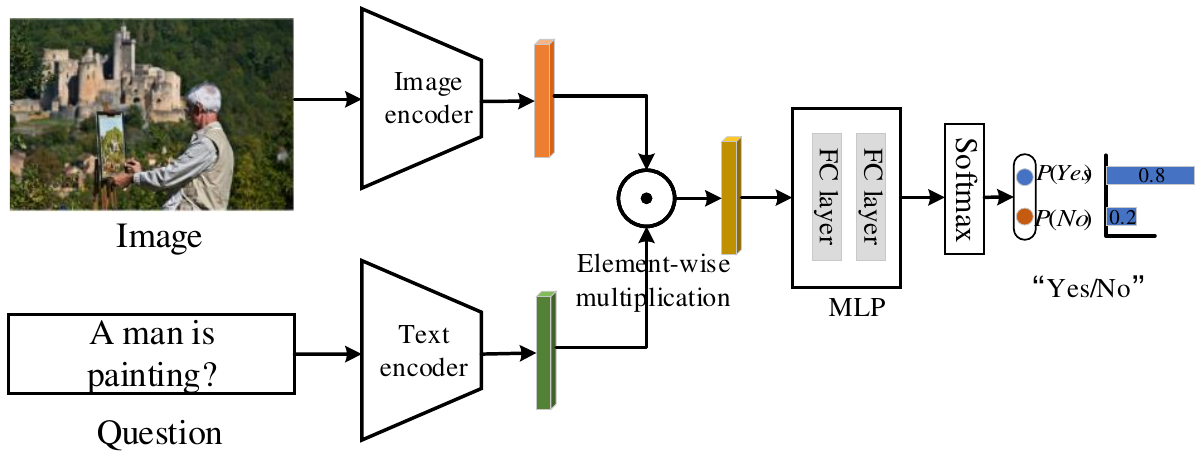}  
 }     
\subfigure[Multi-choice problem] 
{ \label{State_of_the_art_Paris6k}   
  
\includegraphics[width=0.485\columnwidth]{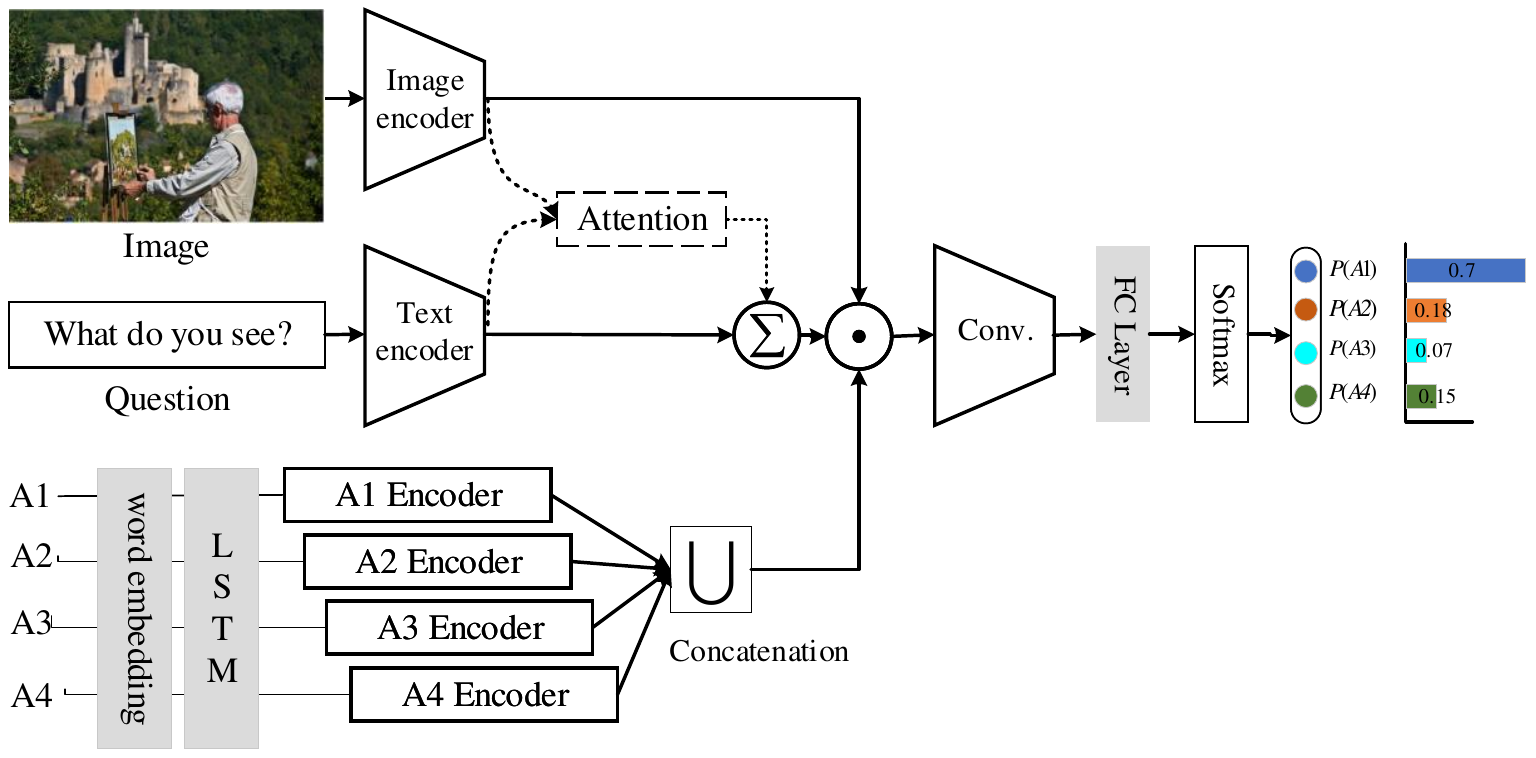}   
  }  
  
 \subfigure[Number counting problem] 
 { \label{State_of_the_art_Oxford5k_1}   
  
\includegraphics[width=0.486\columnwidth]{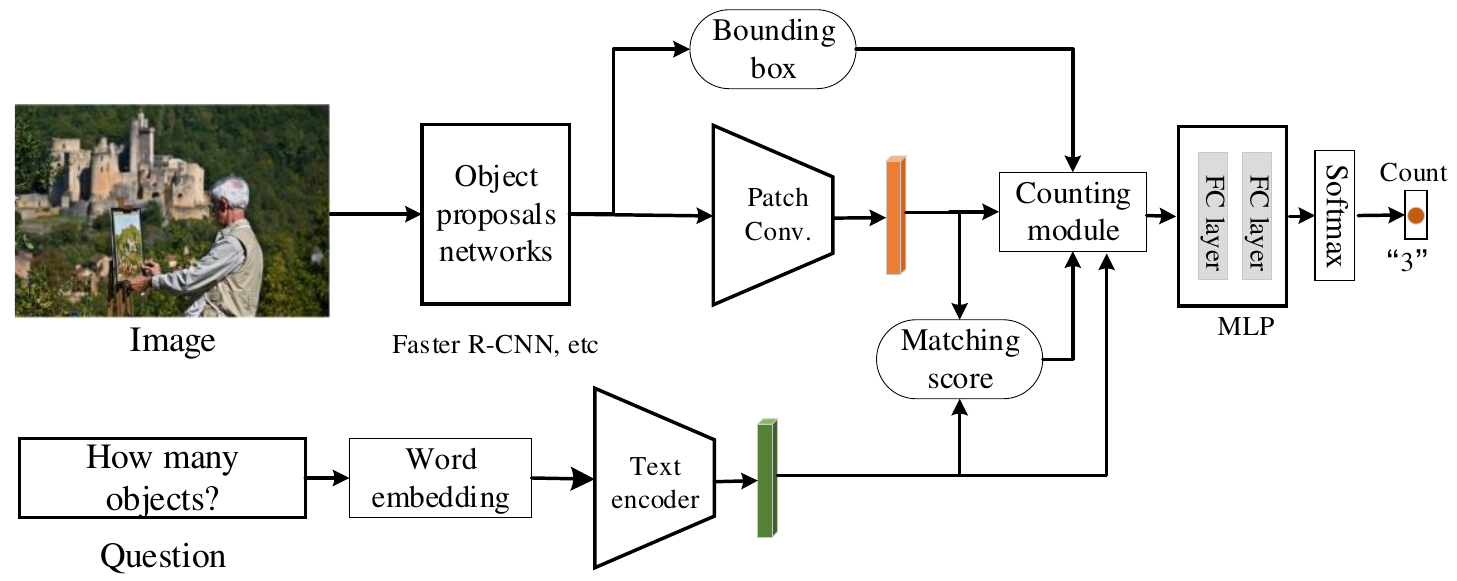}   
  } 
  \subfigure[Open-ended problem] 
{ \label{State_of_the_art_UKBench}   
  
\includegraphics[width=0.485\columnwidth]{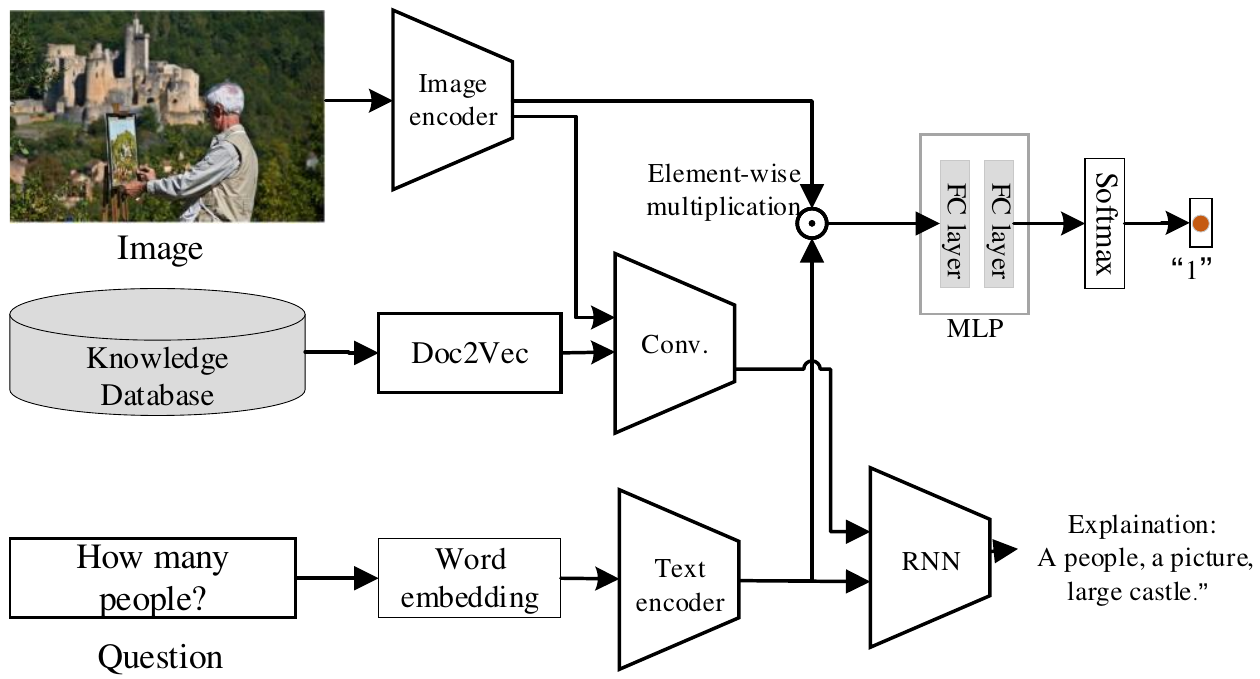}   
  } 
  \vspace{-1em}
\caption{Common types of visual question answering. ``Yes/No'' problem and multi-choice problem can be regarded as a classification problem, while number counting problem and open-ended problem can be viewed as a caption generation problem. }
\label{4_type_VQA}
\end {figure}

Visual question answering (VQA) is a challenging task in which an image and a question are given, then a correct answer is inferred according to visual content and syntactic principle. We summarize four types of VQA \cite{zhang2019information} in Figure \ref{4_type_VQA}. VQA can be categorized into image question answering and video question answering. In this paper, we target recent advances in image question answering. Since VQA was proposed, it has received increasing attention in recent years. For example, there are some training datasets \cite{wu2017visual} built for this task, and some network training tips and tricks are presented in work \cite{teney2018tips}.

To infer correct answers, VQA systems need to understand the semantics and intent of the questions completely, and also should be able to locate and link the relevant image regions with the linguistic information in the questions. VQA applications present two-fold difficulties: feature fusion and reasoning rationality. Thus, VQA more closely reflects the difficulty of multimodal content understanding, which makes VQA applications more difficult than other multimodal applications. Compared to other applications, VQA has various and unknown questions as inputs. Specific details (\emph{e.g.} activity of a person) in the image should be identified along with the undetermined questions. Moreover, the rationality of question answering is based on high-level knowledge and advanced reasoning capability of deep models. As for performance assessment, answers on open-ended questions are difficult to evaluate compared to the other three types in Figure \ref{4_type_VQA} where the answer typically is selected from specific options, or the answer contains only a few words \cite{wu2017visual}.

As summarized in Figure \ref{4_type_VQA}, the research on VQA includes: free-form open-ended questions \cite{Li2018VQAEEE}, where the answer could be words, phrases, and even complete sentences; object counting questions \cite{Zhang2018LearningTC} where the answer is counting the number of objects in one image; multi-choice questions \cite{anderson2018bottom} and Yes/No binary problems \cite{Zhang2016YinAY}. In principle, the type of multi-choice and Yes/No can be viewed as classification problems, where deep models infer the candidate with maximum probability as the correct answer. These two types are associated with different answer vocabularies and are solved by training a multi-class classifier. In contrast, object counting and free-form open-ended questions can be viewed as generation problems \cite{wu2017visual} because the answers are not fixed, only the ones related to visual content and question details.

\textcolor{black}{ Compared to other three mentioned multimodal applications, VQA is more complex and more open-ended. Although much attention has been paid to visual question answering research, there still exist several challenges in this field. One is related to accuracy. Some keywords in question might have been neglected and some visual content might remain unrecognized or misclassified. Because of this, a VQA system might give inaccurate even wrong answers. Another is related to diversity and completeness of the predicted answer, which is especially crucial for free-form open-ended problems, as the output answers should be as complete as possible to explain the given question, and not limited to a specific domain or restricted language forms \cite{wu2017visual}. The third one is the VQA datasets, which should have been less biased. For the existing available datasets, questions that require the use of the image content are often relatively easy to answer.
However, harder questions, such as those beginning with ``Why'', are comparatively rare and difficult to answer since it needs to more reasoning \cite{kafle2017visual}. Therefore, the biased question type impairs the evaluation for VQA algorithms. For recommendations, a larger but less biased VQA dataset is necessary. 
}

\vspace{-1.5em}
\section*{3 Challenges for Deep Multimodal Learning}
\vspace{-1em}

\textcolor{black}{ Typically, domain-specific neural networks process different modalities to obtain individual monomodal representations and are further embedded or aggregated as multimodal features. Importantly, it is still difficult to fully understand how multimodal features are used to perform the aforementioned tasks well. Taking text-to-image generation as an example, we can imagine two questions: First, how can we organize two types of data into a unified framework to extract their features? Second, how can we make sure that the generated image has the same content as the sentence described?}

\begin {figure}[!t]
\captionsetup{font={footnotesize}}
\centering
 { 
   \includegraphics[width=0.9\columnwidth]{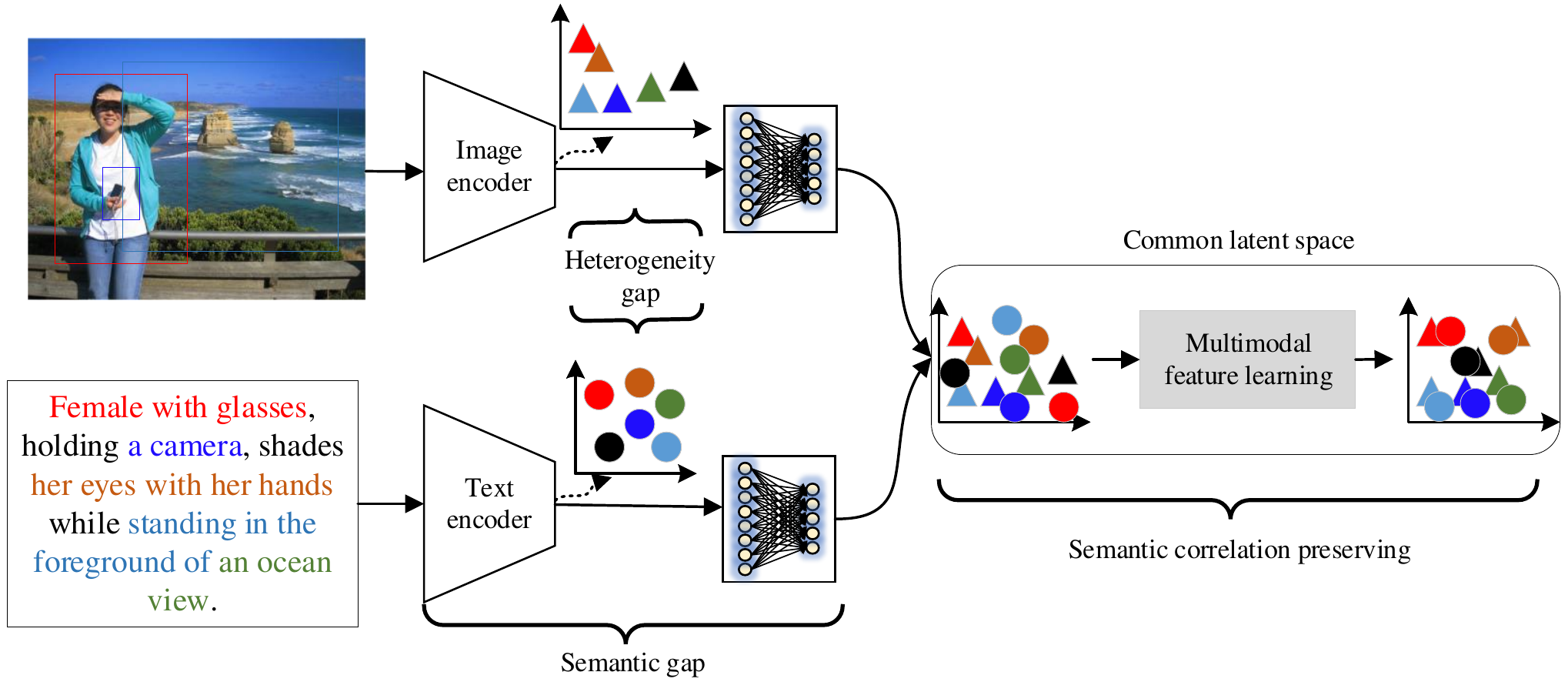} 
 }
 \vspace{-1em}
\caption{ \textcolor{black}{A conceptual illustration of two challenges. We use different shapes to denote different modalities; the circle represents text feature distributions, and the triangle represents image feature distributions. Different shapes with the same color mean that they are semantically similar in content. Apart from the ``semantic gap'' \cite{li2016socializing} which is commonly mitigated in monomodal deep visual tasks such as image classification, the key for deep multimodal content understanding also lies in mitigating the ``heterogeneity gap'' \cite{wang2016comprehensive}, \emph{i.e.} reducing the inter-modality gap and exploring the semantic correlations. }
} 
\label{The_challenge_Multi_modal}
\end {figure}

These two kinds of questions are highly relevant to the heterogeneity gap and the semantic gap in deep multimodal learning. We illustrate the heterogeneity gap and the semantic gap in Figure \ref{The_challenge_Multi_modal}. Recently, much effort has gone into addressing these two challenges. These efforts are categorized into two directions: towards minimizing the heterogeneity gap and towards preserving semantic correlation.

\vspace{-1em}
\subsection*{3.1 Heterogeneity Gap Minimization}
\vspace{-0.5em}

\textcolor{black}{On the one hand, although complementary information from multiple modalities is beneficial for multimodal content understanding, their very different statistical properties can impair the learning of this complementary. For example, an image comprises 3-channel RGB pixel values, whereas the symbolic text consists of words with different lengths. Meanwhile, image and text data contain different ways of conveying semantic information. Usually, the text has more abstract semantics than image while the content of the image is easier and more straightforward to understand than text. }

\textcolor{black}{On the other hand, neural networks are comprised by successive linear layers and non-linear activation functions. The neurons in each layer have different receptive fields so that the systems have various learning capacities. Usually, the last layer of neural networks are used as a way for representing data. Due to the diverse structures of neural networks, the data representations are in various abstractions. Usually, image features are extracted from hierarchical networks and text features are from sequential networks. Naturally, these features are distributed inconsistently so that they are not directly comparable, which leads to the heterogeneity gap. Both the modality data and the network itself contribute to this discrepancy.
}

Therefore, to correlate features among different modalities, it is necessary to construct a common space for these multimodal features to narrow the heterogeneity gap. In general, there are two strategies to narrow the heterogeneity gap. One direction is from the viewpoint of deep multimodal structures and another is from the viewpoint of feature learning algorithms.

Auto-encoders and generative adversarial networks (GANs) \cite{goodfellow2014generative} are two important structures for representing the multimodal structure. We will introduce both of them in the following sections. Generative adversarial networks learn features to bridge image data and text data. For example, GANs are commonly applied to generate images according to their descriptive sentences. \textcolor{black}{This idea is then developed into several variants, such as StackGAN \cite{Han2016StackGAN}, HDGAN \cite{Zhang2018Photographic}, and AttnGAN \cite{Tao18attngan}.} Auto-encoders are used to correlate multimodal features based on feature encoding and feature reconstruction. \textcolor{black}{ For example, Gu et al. \cite{gu2018look}\cite{cao2016correlation} use cross-reconstruction method to preserve multimodal semantic similarity where image (text) features are reconstructed to text (image) features.}

In addition, much effort has gone into minimizing the gaps in uni-modal representations. For instance, sequential neural networks (\emph{e.g.} RNNs) are employed to extract multi-granularity including character-level, word-level, phrase-level and sentence-level text features  \cite{reed2016learning}\cite{li2018read}\cite{you2018end}\cite{chen2018factual}. 
Graph-based approaches have been introduced to explore the semantic relationship in text feature learning \cite{wu2018learning}\cite{jin2018semantic}. 

Regarding the goal of reducing the heterogeneity gap, uni-modal representations are projected into a common latent space under joint or coordinated constraints. Joint representations combine uni-modal features into the same space, while coordinated representations process uni-modal features separately but with certain similarity and structure constraints \cite{baltruvsaitis2019multimodal}. 

\vspace{-1em}
\subsection*{3.2 Semantic Correlation Preserving}
\vspace{-0.5em}

Preserving semantic similarity is challenging. On the one hand, the differences between high-level semantic concepts (\emph{i.e.} features) and low-level values (\emph{e.g.} image pixel) result in a semantic gap among intra-modality embeddings. On the other hand, uni-modal visual and textual representations make it difficult to capture complex correlations across modalities in multimodal learning.

As images and text are used to describe the same content, they should share similar patterns to some extent. \textcolor{black}{Therefore, using several mapping functions, uni-modal representations are projected into the common latent space using individual neural networks (see Figure \ref{The_pipeline_of_deep_multimodal_research}). However, these embedded multimodal features cannot reflect the complex correlations of different modalities because the individual networks are not pre-trained and the feature projection are untargeted. The neural networks mainly act as a way of normalizing the monomodal features into a common latent space.} Therefore, preserving the semantic correlations between projected features of similar image-text pairs is another challenge in deep multimodal research. More specifically, if an image and text are similar in content, the semantic similarity of features in common latent space should be preserved, otherwise, the similarity should be minimized. 

To preserve the semantic correlations, one must measure the similarity between multimodal features where joint representation learning and coordinated representation learning can be adopted. \textcolor{black}{Joint presentation learning is more suitable for the scenarios where all modalities are available during testing stage, such as in visual question answering. For other situations where only one modality is available for testing, such as cross-modal retrieval and image captioning, coordinated representation learning is a better option.}

Generally, feature vectors from two modalities can be concatenated directly in joint representation learning; Then the concatenated features are used to make classification or are fed into a neural network (\emph{e.g.} RNN) for prediction (\emph{e.g.} producing an answer). \textcolor{black}{Simple feature concatenation is a linear operation and less effective, advanced pooling-based methods such as compact bilinear pooling \cite{gao2016compact}\cite{liu2018learning} are introduced to connect the semantically relevant multimodal features.} \textcolor{black}{ Neural networks are also alternative for exploring more corrections on the joint representations. For example, Wang et al. \cite{wang2020dynamic} introduce a multimodal transformer for disentangling the contextual and spatial information so that a unified common latent space for image and text is construct. Similarly, auto-encoders, as as unsupervised structures, are used in several multimodal tasks like cross-modal retrieval \cite{zhan2018comprehensive} and image captioning \cite{zhao2018multi}. The learning capacity of the encoder and decoder are enhanced by improving on the structure of the sub-networks by stacking attention  \cite{yang2016stacked}\cite{fan2018stacked}, paralleling LSTM \cite{zhu2018image}\cite{zhao2018multi}, and ensembling CNN \cite{jiang2018recurrent}. Different sub-networks have their own parameters. Thereby, 
auto-encoders would have more chances to learn comprehensive features. }

 \textcolor{black}{ The key point for coordinated representation learning is to design optimal constraint functions. For example, computing inner product or cosine similarity between two cross-modal features is a simple way to constrain dot-to-dot correlations. Canonical correlation analysis \cite{wu2020augmented}\cite{hotelling1992relations}\cite{Hoshen2018UnsupervisedCA}\cite{tommasi2019combining} is commonly used to maximize semantic correlations between vectors; For better performance and stability improvement, metric learning methods such as bi-directional objective functions \cite{Liu2017LearningAR}\cite{zheng2017dual}\cite{wang2019learning} are utilized. However, mining useful samples and selecting appropriate margin settings remain empirical in metric learning \cite{zhang2018deep}. Regarding these limits in metric learning, some new methods, such as adversarial learning \cite{wang2017adversarial}\cite{li2018self}\cite{peng2019cm} and KL-divergence \cite{zhang2018deep}, are introduced for these demerits. Instead of selecting three-tuple samples, these alternative methods consider the whole feature distributions in a common latent space. In addition, attention mechanisms \cite{Wang2019HierarchicalAttention}\cite{Song2019connectingLanguage}\cite{Zhou2019Dynamic} and reinforcement learning \cite{ren2017deep}\cite{Wang2019ADR} are popularly employed to align relevant features between modalities.}
 
To address the above-mentioned challenges, several new ideas, including methods for feature extraction, structures of deep networks, and approaches for multimodal feature learning, have been proposed in recent years. The advances from these ideas are introduced in the following sections.

\vspace{-1.5em}
\section*{4 Recent Advances in Deep Multimodal Feature Learning}
\vspace{-1em}

\begin {figure*}[!t]
\captionsetup{font={footnotesize}}
\centering
 { 
   \includegraphics[width=0.9 \columnwidth]{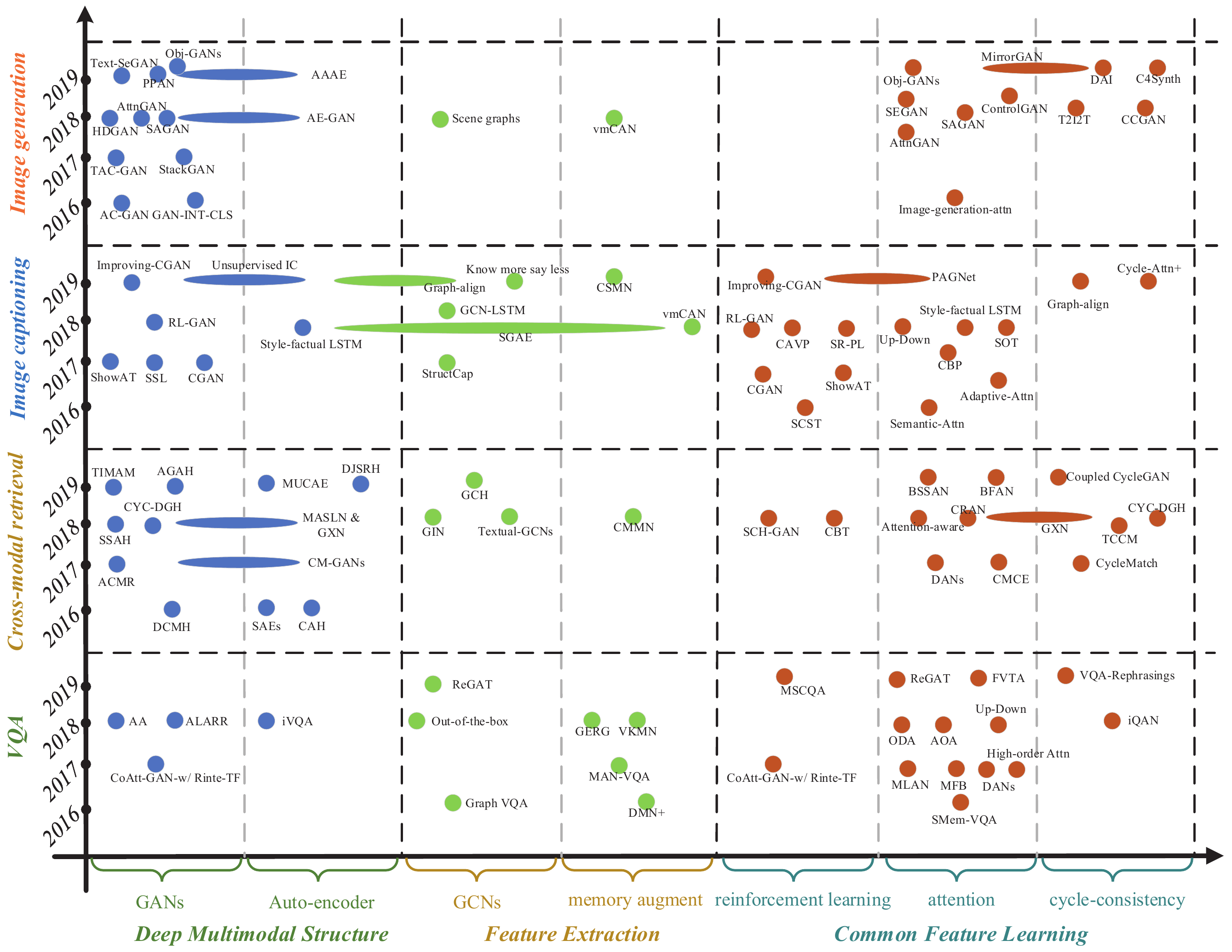} 
 }
 \vspace{-1em}
\caption{ Representative approaches for multi-modal content understanding. We categorize these new ideas and trends in the perspective of deep multimodal structure, feature extraction and common feature learning, which are applied into different applications: Text-SeGAN \cite{cha2019adversarial}, PPAN \cite{Gao2019Perceptual}, MUCAE \cite{liu2019cross}, AAAE \cite{xu2019adversarially}, HDGAN \cite{Zhang2018Photographic}, AttnGAN \cite{Tao18attngan}, SAGAN \cite{zhang2019self}, AE-GAN \cite{wang2018unregularized}, TAC-GAN \cite{dash2017tac}, StackGAN \cite{Han2016StackGAN}, AC-GAN \cite{odena2017conditional}, GAN-INT-CLS \cite{reed2016generative}, Unsupervised IC \cite{feng2019unsupervised}, Improving-CGAN \cite{chen2019improving}, RL-GAN \cite{yan2018image}, ShowAT \cite{chen2017show}, SSL \cite{shetty2017speaking}, CGAN \cite{dai2017towards}, SSAH \cite{li2018self}, CYC-DGH \cite{wu2019cycle}, MASLN \cite{xu2018modal}, GXN \cite{gu2018look}, ACMR \cite{wang2017adversarial}, CM-GANs \cite{Peng2017CMGANsCG}, DCMH \cite{Jiang2017DeepCH}, TIMAM \cite{sarafianos2019adversarial}, AGAH \cite{gu2019adversary}, DJSRH \cite{su2019deep}, SAEs \cite{wang2016effective}, CAH \cite{cao2016correlation}, AA \cite{sharma2018attend}, ALARR \cite{liu2018adversarial}, iVQA \cite{liu2018inverse}, CoAtt-GAN-w/Rinte-TF \cite{wu2018you}, Scene graphs \cite{Johnson2018Image}, vmCAN \cite{zhang2018text}, Graph-align \cite{gu2019unpaired}, Know more say less \cite{li2019know}, GCN-LSTM \cite{Yao2018ExploringVR}, SGAE \cite{Yang2018AutoEncodingSG}, StructCap \cite{Chen2017StructCapSS}, GCH \cite{xu2019graph}, GIN \cite{yu2018modeling}, Textual-GCNs \cite{yu2019semantic}, CSMN \cite{park2018towards}, CMMN \cite{song2018deep}, ReGAT \cite{Li2019RelationawareGA}, Out-of-the-box \cite{narasimhan2018out}, Graph VQA \cite{teney2017graph}, GERG \cite{Li2018TextbookQA}, VKMN \cite{Su2018LearningVK}, MAN-VQA \cite{Ma2018VisualQA}, DMN+ \cite{Xiong2016DynamicMN}, MSCQA \cite{Wang2019ADR}, SCH-GAN \cite{zhang2018sch}, CBT \cite{qi2018crossmodal}, SCST \cite{rennie2017self}, CAVP \cite{liu2018context}, SR-PL \cite{liu2018show}, SMem-VQA \cite{Xu2016AskAA}, ODA \cite{wu2018object}, AOA \cite{singh2018attention}, Up-Down \cite{anderson2018bottom}, Attention-aware \cite{zhang2018attention}, BSSAN \cite{huang2019bi}, CRAN \cite{qi2018cross}, CBP \cite{cui2018learning}, SOT \cite{chen2018show}, PAGNet \cite{Song2019connectingLanguage}, MirrorGAN \cite{qiao2019mirrorgan}, DAI \cite{lao2019dual}, T2I2T \cite{Gorti2018Text}, CCGAN \cite{Liu2018SemanticIS}, C4Synth \cite{joseph2019c4synth}, Cycle-Attn+ \cite{wu2019improving}, Coupled CycleGAN \cite{li2019coupled}, TCCM \cite{cornia2018towards}, CycleMatch \cite{liu2019cyclematch}, VQA-Rephrasings \cite{shah2019cycle}, iQAN \cite{li2018visual}, MLAN \cite{yu2017multilevel}, MFB \cite{yu2017multi}, DANs \cite{nam2017dual}, High-order Attn \cite{schwartz2017high}, FVTA \cite{liang2019focal}, CMCE \cite{li2017identity}, BFAN \cite{liu2019focus}, Semantic-Attn\cite{you2016image}, Adaptive-Attn \cite{lu2017knowingtolook}, Obj-GANs \cite{li2019object}, Image-generation-attn \cite{mansimov2015generating}, ControlGAN \cite{li2019controllable}, SEGAN \cite{tan2019semantics}.  }
\label{Methods_Application}
\end {figure*}

\textcolor{black}{ Regarding the aforementioned challenges, exploring content understanding between image and text has attracted sustained attention and lots of remarkable progresses have been made. In general, these advances are mainly from a viewpoint of network structure and a viewpoint of feature extraction/enhancement. To this end, combining the natural process pipeline of multimodal research (see Figure \ref{The_pipeline_of_deep_multimodal_research}), we categorize these research ideas into three groups: deep multimodal structures presented in Section 4.1, multimodal feature extraction approaches introduced in Section 4.2, and common latent space learning described in Section 4.3. Deep multimodal structures indicate the basic framework in community; Multimodal feature extraction is the prerequisite which supports the following similarity exploring; Common latent space learning is the last but a critical procedure to make the multimodal features comparable. For a general overview for these aspects in multimodal applications, we make a chart for the representative methods in Figure \ref{Methods_Application}.}

\vspace{-1em}
\subsection*{4.1 Deep Multimodal Structures}
\vspace{-0.5em}

\textcolor{black}{ Deep multimodal structures are the fundamental frameworks to support different deep networks for exploring visual-textual semantics. These frameworks, to some extent, have critical influences for the following feature learning steps (\emph{i.e.} feature extraction or common latent space learning). To understand the semantics between images and text, deep multimodal structures usually involve computer vision and natural language processing (NLP) field \cite{young2018recent}.  For instance, raw images are processed by hierarchical networks such as CNNs and raw input text can be encoded by sequential networks such as RNN, LSTM \cite{chen2018factual}, and GRU \cite{young2018recent}. During the past years, a variety of related methods have blossomed and accelerated the performance of multimodal learning directly in multimodal applications, as shown in Figure \ref{Methods_Application}. }

Deep multimodal structures include generative models, discriminative models. Generative models implicitly or explicitly represent data distributions measured by a joint probability $ P (X, Y)$, where both raw data $ X $ and ground-truth labels $ Y $ are available in supervised scenarios. Discriminative models learn classification boundaries between two different distributions indicated by conditional probability $ P (Y\!\! \mid \!\! X)$. Recent representative network structures for multimodal feature learning are auto-encoders and generative adversarial networks. There are some novel works to improve the performance of multimodal research based on these two basic structures (see Figure \ref{Methods_Application}).
 
 \vspace{-1em}
\subsubsection*{4.1.1 Auto-encoders}
 \vspace{-0.5em}

\begin {figure}[!t]
\centering
 \subfigure[Multimodal autoencoder]
 { \label{Multimodal_autoencoder}     
   \includegraphics[width=0.46\columnwidth]{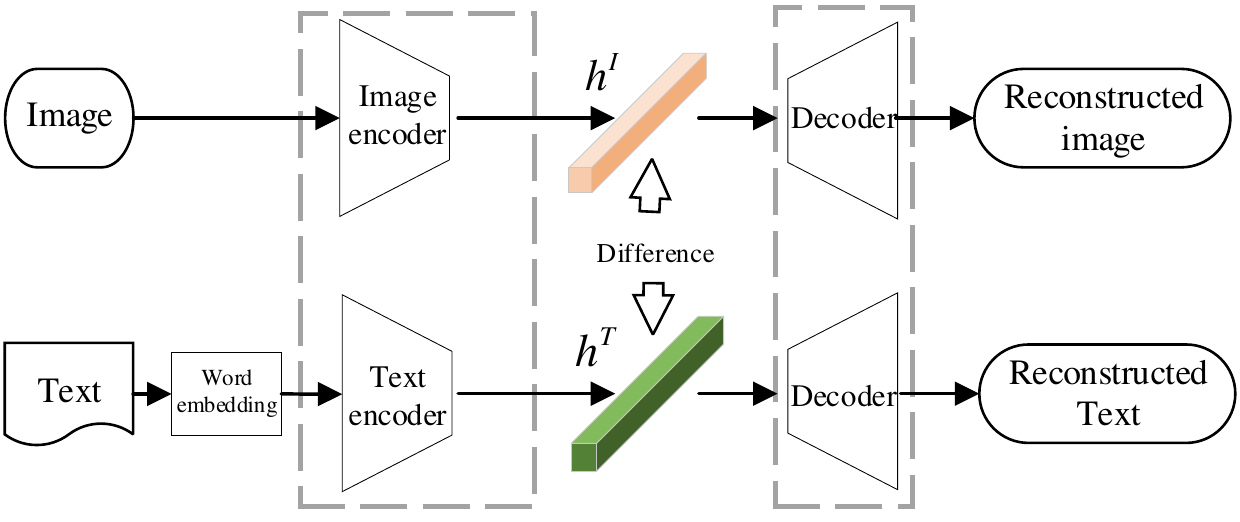}  
 }     
\subfigure[Bimodal autoencoder] 
{ \label{Bimodal_autoencoder}   
  
\includegraphics[width=0.51\columnwidth]{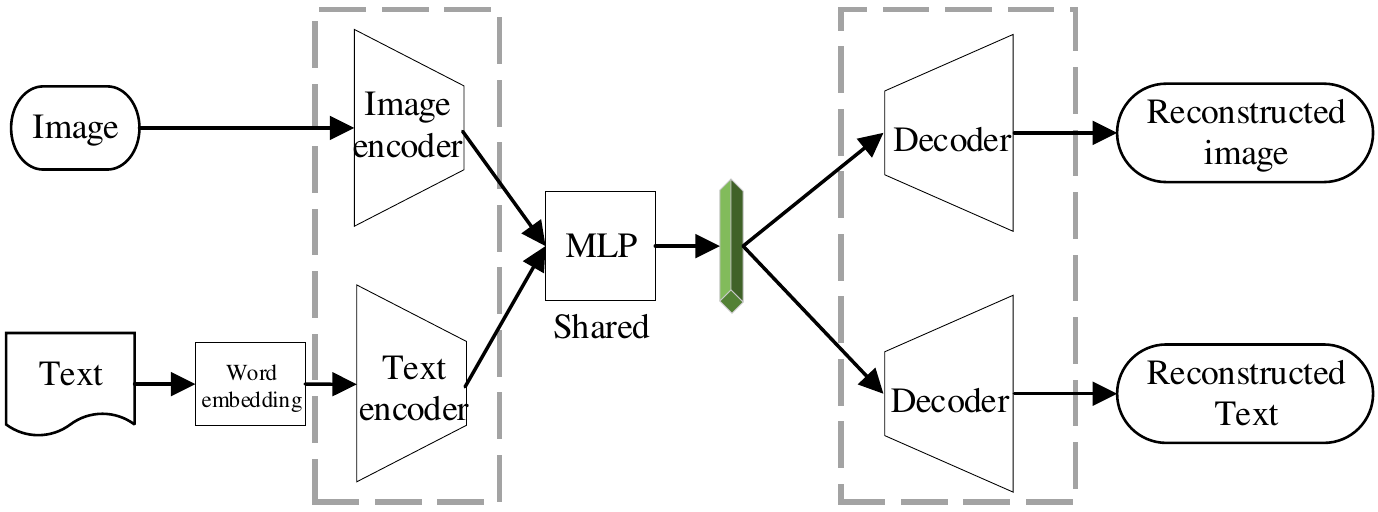}   
  }  
  
  \subfigure[Correlation autoencoder] 
 { \label{Correction_autoencoder}   
  
\includegraphics[width=0.48\columnwidth]{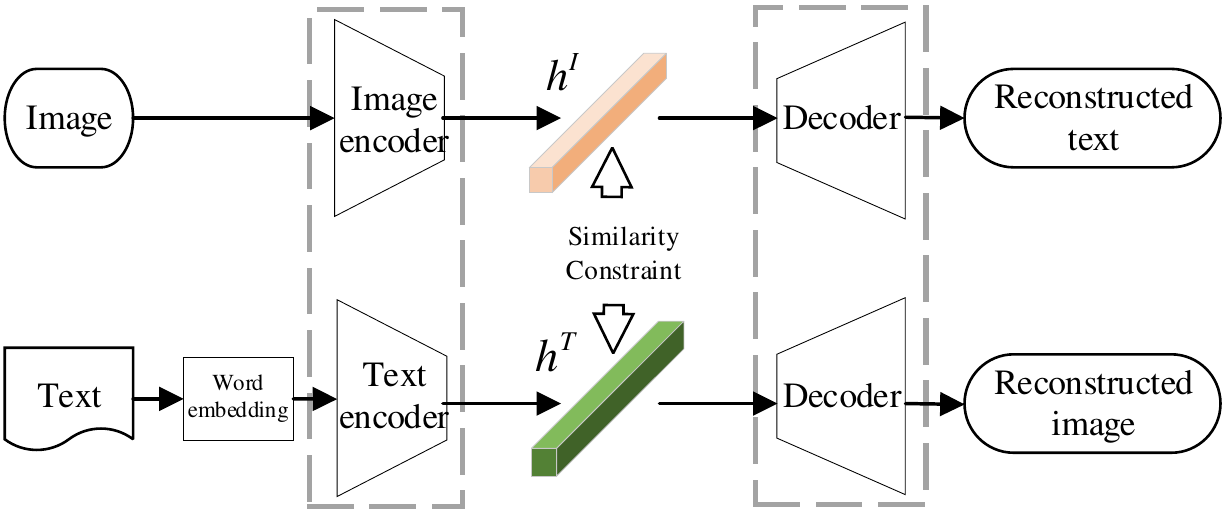}   
  }
  \vspace{-1em}
\caption{ Convolutional autoencoder used for deep multimodal learning. The branch for image feature learning can adopt hierarchical networks such as CNNs; the branch for text feature learning can capture the dependency relations in a sentence by sequential models such as RNN and LSTM. Usually, a reconstruction loss function is used to optimize network training.}
\label{AutoEncoder}
\end {figure}

\textcolor{black}{The main idea of auto-encoder for multimodal learning is to first encode data from a source modality as hidden representations and then to use a decoder to generate features (or data) for the target modality. Thus, it is commonly used in dimensionality reduction for bi-directional applications where two modalities are available at the test stage.}
For this structure, reconstruction loss is the constraint for training encoder and decoder to well capture the semantic correlations between image and text features. For clarity, we identify three ways for correlation learning using auto-encoders in Figure \ref{AutoEncoder}. For instance, as shown in Figure \ref{Multimodal_autoencoder}, the input images and text are processed separately with non-shared encoder and decoder, after which these hidden representations from the encoder are coordinated through a constraint such as Euclidean distance \cite{liu2019cross}. The coordinated methods can be replaced by joint methods in Figure \ref{Bimodal_autoencoder} where image and text features are projected into a common space with a shared multilayer perceptron (MLP). Subsequently, the joint representation is reconstructed back to the original raw data \cite{xu2018modal}. Alternatively, feature correlations are captured by cross reconstruction with similarity constraints (\emph{e.g.} a similarity matrix is used as supervisory information \cite{Jiang2017DeepCH}\cite{cao2016correlation}) between hidden features. \textcolor{black}{ The idea of constraining sample similarity is also incorporated with GANs into a cycle-consistent formation for cross-modal retrieval like GXN \cite{gu2018look} and CYC-DGH\cite{wu2019cycle} (see Figure \ref{Methods_Application}).}

\textcolor{black}{The neural networks contain in the encoder-decoder framework can be modality specific. For image data, the commonly used neural networks are CNN while sequential networks like LSTM are most often used for text data. When applied for multimodal learning, the decoder (\emph{e.g.} LSTM) constructs hidden representations of one modality in another modality. The goal is not to reduce reconstruction error but to minimize the output likelihood estimation. Therefore, most works focus on the decoding since it is a process to project the less meaningful vectorial representations to meaningful outputs in target modality.} Under this idea, several extensions have been introduced. \textcolor{black}{ The main difference among these algorithms lies in the structure of the decoder. For example, ``stack and parallel LSTM'' \cite{zhu2018image}\cite{zhao2018multi} is to parallelize more parameters of LSTMs to capture more context information. Similar ideas can be found in ``CNN ensemble learning'' \cite{jiang2018recurrent}. Instead of grabbing more information by stacking and paralleling, ``Attention-LSTM'' \cite{zhu2018image}\cite{jiang2018learning} combines attention technique into LSTM to highlight most relevant correlations, which would be more targeted. An adversarial training strategy is employed into the decoder to make all the representations discriminative for semantics but indiscriminative for modalities so that intra-modal semantic consistency is effectively enhanced \cite{xu2018modal}. \textcolor{black}{Considering the fixed structure in the decoder like RNN might limit the performance, Wang et al. \cite{wang2020evolutionary} introduce evolutionary algorithm to adaptively generate neural network structures in the decoder. }}

\vspace{-1em}
\subsubsection*{4.1.2 Generative Adversarial Networks}
\vspace{-0.5em}

As depicted in Figure \ref{Methods_Application}, adversarial learning from generative adversarial networks \cite{goodfellow2014generative} has been employed into applications including image captioning \cite{shetty2017speaking}\cite{chen2019improving}\cite{chen2017show}, cross-modal retrieval \cite{wang2017adversarial}\cite{li2018self}\cite{peng2019cm}\cite{xu2018modal}
\cite{zhang2018sch}\cite{zhang2018attention} and image generation \cite{cha2019adversarial}\cite{Zhang2018Photographic}\cite{Han2016StackGAN}\cite{HanZ2017StackGAN}\cite{Gao2019Perceptual}\cite{Tao18attngan}, but has been less popular in VQA tasks. GANs combine generative sub-models and discriminative sub-models into a unified framework in which two components are trained in an adversary manner. 

Different from auto-encoders, GANs can cope with the scenarios where there are some missing data. To accurately explore the correlations between two modalities, multimodal research works involving GANs have been focusing on the whole network structure and its two components: \textit{generator} and \textit{discriminator}.

\textcolor{black}{ For the generator which also can be viewed as an encoder, an attention mechanism is often used to capture the important key points and align cross-modal features such as AttnGAN \cite{Tao18attngan} and Attention-aware methods \cite{zhang2018attention}. Sometimes, Gaussian noise is concatenated with the generator's input vector to improve the diversity of generated samples and avoid model collapse, such as the conditioning augmentation block in StackGAN \cite{Han2016StackGAN}. To improve its capacity for learning hierarchical features, a generator can be organized into different nested structures such as hierarchical-nested \cite{Zhang2018Photographic} and hierarchical-pyramid \cite{Gao2019Perceptual}, both of them can capture multi-level semantics.}

 \textcolor{black}{ The discriminator, which usually performs binary classification, attempts to discriminate the ground-truth labels from the outputs of the generator. Some recent ideas are proposed to improve the discrimination of GANs. Originally, discriminator in the first work \cite{goodfellow2014generative} just needs to classify different distributions into ``\textit{True}'' or ``\textit{False}'' \cite{reed2016generative}. However, discriminator can also make a class label classification where a label classifier is added on the top of discriminator \cite{odena2017conditional}\cite{dash2017tac}.
Apart from the label classification, a semantic classifier is designed to further predict semantic relevances between a synthesized image and a ground-truth image for text-to-image generation \cite{cha2019adversarial}. Only focusing on the paired samples leads to relatively-weak robustness. Therefore, the unmatched image-text samples can be fed into a discriminator (\emph{e.g.} GAN-INT-CLS \cite{reed2016generative} and AACR \cite{wu2020augmented}) so that the discriminator would have a more powerful discriminative capability.}

According to previous work \cite{He2018An}, the whole structure of GANs in multimodal research are categorized into direct methods \cite{reed2016generative}\cite{dash2017tac}\cite{odena2017conditional}, hierarchical methods \cite{Zhang2018Photographic}\cite{Gao2019Perceptual} and iterative methods \cite{Han2016StackGAN}\cite{HanZ2017StackGAN}\cite{Tao18attngan}. Contrary to direct methods, hierarchical methods divide raw data in one modality (\emph{e.g.} image) into different parts such as a ``style'' and ``structure'' stage, thereby, each part is learned separately. Alternatively, iterative methods separate the training into a ``coarse-fine'' process where details of the results from a previous generator are refined. Besides, cycle-consistency from cycleGAN \cite{gu2019unpaired} is introduced for unsupervised image translation where a self-consistency (reconstruction) loss tries to retain the patterns of input data after a cycle of feature transformation. This network structure is then applied into tasks like image generation \cite{Liu2018SemanticIS}\cite{Gorti2018Text} and cross-modal retrieval \cite{gu2018look}\cite{wu2019cycle} to learn semantic correlation in an unsupervised way.

Preserving semantic correlations between two modalities is to reduce the difference of inconsistently distributed features from each modality. Adversarial learning keeps pace with this goal. In recent years, adversarial learning is widely used to design algorithms for deep multimodal learning \cite{peng2019cm}\cite{wang2017adversarial}\cite{zhang2018attention}\cite{li2018self}\cite{zhang2018sch}\cite{xu2018modal}. For these algorithms, there are no classifiers for binary classification. Instead, two sub-networks are trained with the constraints of competitive loss functions.

As the dominant popularity of adversarial learning, some works are performed by combining auto-encoders and GANs in which the encoder in auto-encoders and the generator in GANs share the same sub-network \cite{xu2019adversarially}\cite{wang2018unregularized}\cite{feng2019unsupervised}\cite{xu2018modal}\cite{gu2018look}\cite{Peng2017CMGANsCG} (see Figure \ref{Methods_Application}). For example, in the first work about unsupervised image captioning \cite{feng2019unsupervised}, the core idea of GANs is used to generate meaningful text features from scratch of text corpus and cross-reconstruction is performed between synthesized text features and true image features.

\vspace{-1em}
\subsection*{4.2 Multimodal Feature Extraction}
\vspace{-0.5em}

\textcolor{black}{ Deep multimodal structures support the following learning process. Thereby, feature extraction is closer for exploring visual-textual content relations, which is the prerequisite to discriminate the complementarity and redundancy of multiple modalities. It is well-known that image features and text features from different deep models have distinct distributions although they convey the same semantic concept, which results in a heterogeneity gap. In this section, we introduce several effective multimodal feature extraction methods for addressing the heterogeneity gap. In general, these methods focus on (1) learning the structural dependency information to reasoning capability of deep neural networks and (2) storing more information for semantic correlation learning during model execution. Moreover, (3) feature alignment schemes using attention mechanism are also widely explored for preserving semantic correlations.}

\subsubsection*{4.2.1 Graph Embeddings with Graph Convolutional Networks}
\vspace{-0.5em}

\textcolor{black}{Words in a sentence or objects within an image have some dependency relationships, and graph-based visual relationship modelling is beneficial for the characteristic 
\cite{wang2020learning}.} Graph Convolutional Networks (GCNs) are alternative neural networks designed to capture this dependency information. Compared to standard neural networks such as CNNs and RNNs, GCNs would build a graph structure which models a set of objects (nodes) and their dependency relationships (edges) in an image or sentence, embed this graph into a vectorial representation, which is subsequently integrated seamlessly into the follow-up steps for processing. Graph representations reflect the complexity of sentence structure and are applied to natural language processing such as text classification \cite{kipf2016semi}. For deep multimodal learning, GCNs receive increasing attention and have achieved breakthrough performance on several applications, including cross-modal retrieval \cite{Yu2018TextualRM}, image captioning \cite{Yao2018ExploringVR}\cite{Yang2018AutoEncodingSG}\cite{wang2020learning}\cite{Chen2017StructCapSS}, and VQA \cite{teney2017graph}\cite{narasimhan2018out}\cite{Li2019RelationawareGA}. Recent reviews \cite{Zhou2018GraphNN}\cite{Wu2019ACS} have reported comprehensive introductions to GCNs. However, we focus on recent ideas and processes in deep multimodal learning.

Graph convolutional networks in multimodal learning can be employed in text feature extraction \cite{Yu2018TextualRM}\cite{teney2017graph}\\
\cite{wang2020learning}\cite{yu2020learning} and image feature extraction \cite{Yao2018ExploringVR}\cite{Yang2018AutoEncodingSG}\cite{Chen2017StructCapSS}. Among these methods, GCNs capture semantic relevances of intra-modality according to the neighborhood structure. GCNs also capture correlations between two modalities according to supervisory information. Note that vector representations from graph convolutional networks are fed into subsequent networks (\emph{e.g.} ``encoder-decoder'' framework) for further learning.

\textcolor{black}{ GCNs aim at determining the attributes of objects and subsequently characterize their relationships. On the one hand, GCNs can be applied in a singular modality to reduce the intra-modality gap. For instance, Yu et al. \cite{Yu2018TextualRM} introduce a ``GCN+CNN'' architecture for text feature learning and cross-modal semantic correlation modeling. In their work, Word2Vec and \textit{k}-nearest neighbor algorithm are utilized to construct semantic graphs on text features. GCNs are also explored for image feature extractions, such as in image captioning \cite{Yao2018ExploringVR}\cite{Yang2018AutoEncodingSG}\cite{Chen2017StructCapSS}. In previous work \cite{Chen2017StructCapSS}, a tree structure embedding scheme is proposed for semantic graph construction. Specifically, input images are parsed into several key entities and their relations are organized into a visual parsing tree (VP-Tree). This process can be regarded as an encoder. The VP-Tree is transformed into an attention module to participate in each state of LSTM-based decoder.
VP-Tree based graph construction is somewhat in a unified way. Alternative methods are introduced to construct more fine-graded semantic graphs \cite{Yao2018ExploringVR}\cite{Yang2018AutoEncodingSG}. Specifically, object detectors (\emph{e.g.} Faster-RCNN \cite{Ren2015FasterRT}) and visual relationship detectors (\emph{e.g.} MOTIFS \cite{Zellers2018NeuralMS}) are used to get image regions and spatial relations, semantic graphs and spatial graphs are constructed based on the detected regions and relations, respectively. Afterwards, GCNs extract visual representations based on the built semantic graphs and spatial graphs.}

\textcolor{black}{
Graph convolutional networks are also introduced to mitigate the inter-modality gap between image and text \cite{teney2017graph}\cite{narasimhan2018out}. Take the work \cite{narasimhan2018out} for VQA as an example, an image is parsed into different objects, scenes, and actions. Also, a corresponding question is parsed and processed to obtain its question embeddings and entity embeddings. These embedded vectors of image and question are concatenated into node embeddings then fed into graph convolutional networks for semantic correlation learning. Finally, the output activations from graph convolutional networks are fed into sequential networks to predict answers.}

\textcolor{black}{ As an alternative method, graph convolutional networks are worthy more exploration for correlations between two modalities. Moreover, there exist two limitations in graph convolutional networks. On the one hand, graph construction process is overall time- and space-consuming; On the other hand, the accuracy of output activations from graph convolutional networks mostly relies on supervisory information to construct an adjacency matrix by training, which are more suitable for structured data, so flexible graph embeddings for image and/or text remains an open problem.}

\vspace{-1em}
\textcolor{black}{\subsubsection*{4.2.2 Memory-augmented Networks}}
\vspace{-0.5em}

\textcolor{black}{To enable deep networks to understand multimodal content and have better reasoning capability for various tasks, a solution may be the mentioned GCNs. Moreover, another solution that has gained attention recently is memory-augmented networks. Directly, when much information in mini-batch even the whole dataset is stored in a memory bank, such networks have greater  capacity to memorize correlations.}

In conventional neural networks like RNNs for sequential data learning, the dependency relations between samples are captured by the internal memory of recurrent operations. However, these recurrent operations might be inefficient in understanding and reasoning overextended contexts or complex images. For instance, most captioning models are equipped with RNN-based encoders, which predict a word at every time step based only on the current input and hidden states used as implicit summaries of previous histories. However, RNNs and their variants often fail to capture long-term dependencies \cite{park2018towards}. For this limitation, memory networks \cite{sukhbaatar2015end} are introduced to augment the memory primarily used for text question-answering \cite{zhang2019information}. Memory networks improve understanding of both image and text, and then ``remember'' temporally distant information.

\begin {figure}[!t]
\centering
 { 
   \includegraphics[width=0.8\columnwidth]{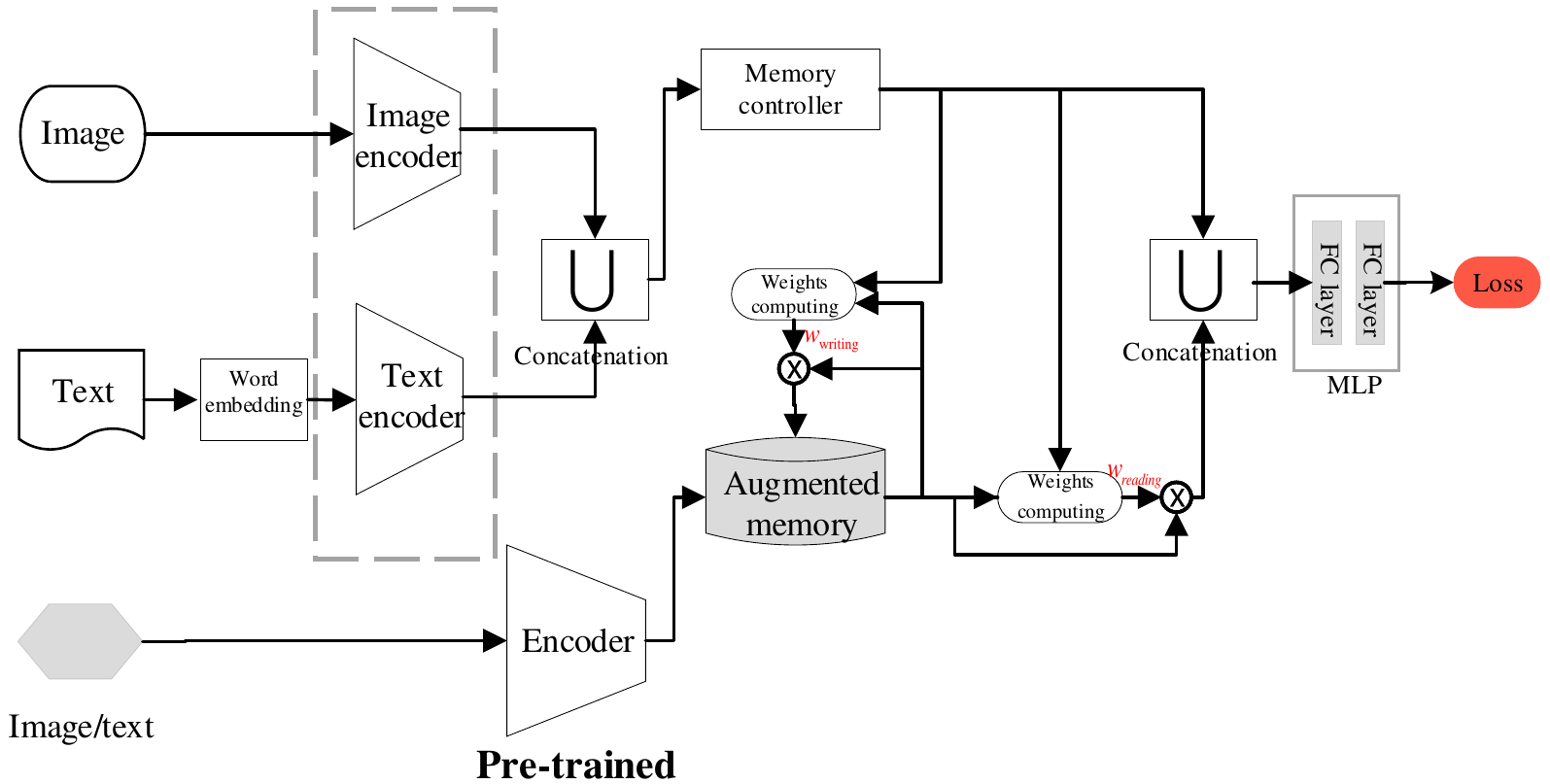} 
 }
 \vspace{-1em}
\caption{ The paradigm of memory-augmented networks in multimodal learning. Augmented memory stores the features of all long-term sequences processed by a pre-trained classifier. The stored vectors can be selected (read) by the memory controller according to the reading weights. Meanwhile, these stored vectors are updated according to the writing weights at each time-step. }
\label{pipe_deep_memory_network}
\end {figure}

\textcolor{black}{Memory-augmented networks can be regarded as recurrent neural networks with explicit attention methods that select certain parts of the information to store in their memory slots. As reported in Figure \ref{Methods_Application}, memory-augmented networks are used in cross-modal retrieval \cite{song2018deep}, image captioning \cite{park2018towards}\cite{chen2018factual}\cite{Park2017AttendTY}\cite{wu2018image} and VQA \cite{Su2018LearningVK}\cite{Ma2018VisualQA}\cite{Xiong2016DynamicMN}\cite{Li2018TextbookQA}\cite{Xu2016AskAA}. We illustrate memory-augmented networks for multimodal learning in Figure \ref{pipe_deep_memory_network}. A memory block, which acts as a compressor, encodes the input sequence into its memory slots. The memory slots are a kind of external memory to support learning; the row vectors in each slot are accessed and updated at each time-step. During training, a network such as LSTM or GRU, which acts as a memory controller, refers to these memory slots to compute reading weights (see Figure \ref{pipe_deep_memory_network}). According to the weights, the essential information is obtained to predict the output sequence. Meanwhile, the controller computes writing weights to update values in memory slots for the next time-step of the training \cite{hudson2018compositional}.}

 \textcolor{black}{ Memory networks can expand the ``memory'' of networks thus store more information. The network performance relates to the memory slots' initialization strategy and the stored information. For this aspect, memory networks have been combined with other techniques like attention mechanisms \cite{fan2018stacked} to further improve its feature learning capability. For example, Xiong et al. \cite{Xiong2016DynamicMN} explore the impact of different initialization strategies to demonstrate that initializations from the outputs of pre-trained networks have better performance. This was verified in works \cite{Xu2016AskAA} where output features from image patches are stored into memory slots of spatial memory networks for VQA. Thereby, generated answers are updated based on gathering evidence from the accessed regions in memory slots. Similarly, Ma et al. \cite{Ma2018VisualQA} adopt LSTM to obtain text features of each sentence and store into memory slots. Then memory-augmented networks are utilized to determine the importance of concatenated visual and text features over the whole training data. Further considering both two modalities, a visual knowledge memory networks is introduced in which memory slots store key-value vectors computed from images, query questions and a knowledge base \cite{Su2018LearningVK}; 
 Instead of storing the actual output features, Song et al. \cite{song2018deep} adopt memory slots to store a prototype concept representation from pre-trained concept classifiers, which is inspired from the process of human memory.}

Memory-augmented networks improve the performance of deep multimodal content understanding by offering more information to select. However, this technique is less popular in image generation, image captioning and cross-modal retrieval than VQA (see in Figure \ref{Methods_Application}). A possible reason for this is that in cross-modal retrieval, memory-augmented networks might require extra time when memory controllers determine when to write or read from the external memory blocks. It will hurt overall retrieval efficiency.

\vspace{-0em}
\subsubsection*{ 4.2.3 Attention Mechanism}
\vspace{-0em}

\textcolor{black}{ As mentioned in Section 3.2, one challenge for deep multimodal learning is to preserve semantic correlations among multimodal features. Regarding this challenge for content understanding, feature alignment plays a crucial role. Image and text features are first processed by deep neural networks under a certain structure like auto-encoders. Naturally, the final output global features include some irrelevant or noisy background information, which is not optimal for performing multimodal tasks.}

\begin {figure}[!t]
\centering
 \subfigure[Visual attention]
 { \label{Visual_attention}     
   \includegraphics[width=0.48\columnwidth]{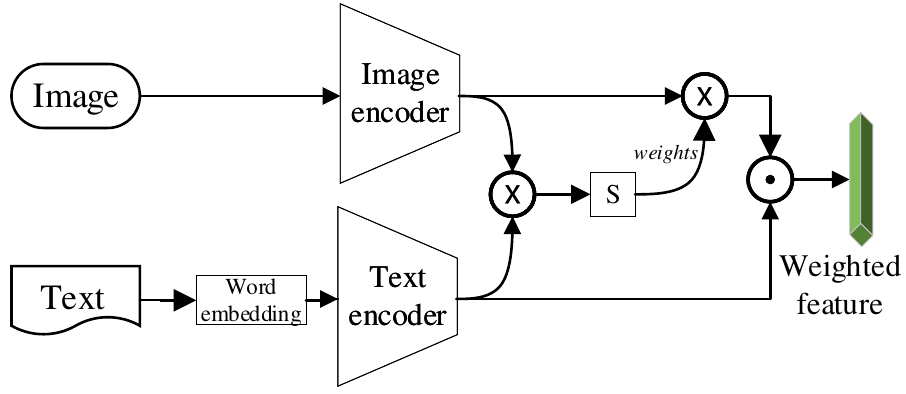}  
 }     
\subfigure[Textual attention] 
{ \label{Textual_attention}   
  
\includegraphics[width=0.48\columnwidth]{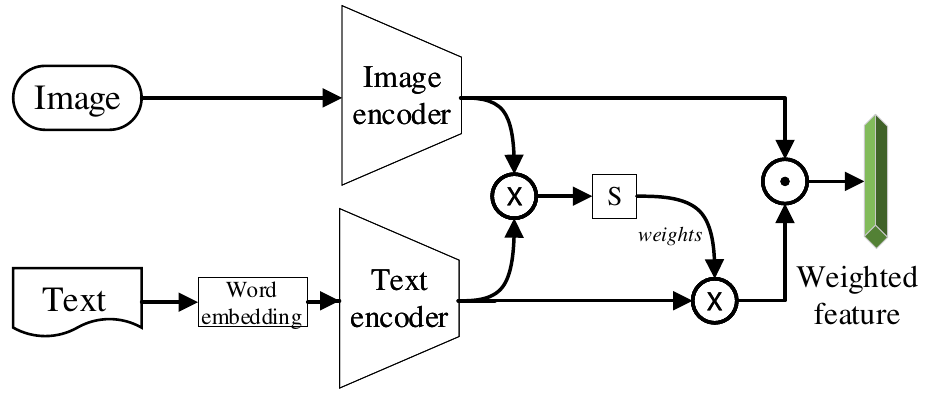}   
  }  
  
  \subfigure[Co-attention] 
 { \label{Co_attention}   
  
\includegraphics[width=0.48\columnwidth]{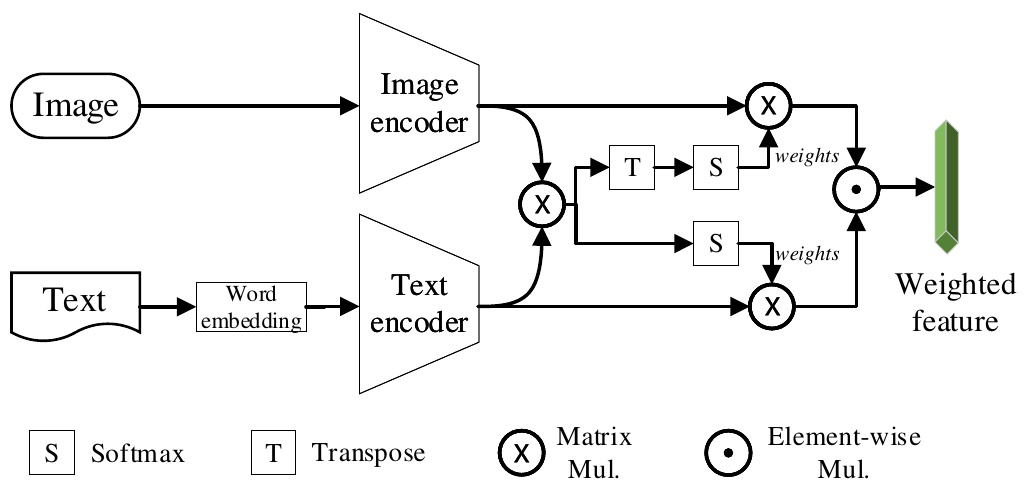}   
  }
  \subfigure[Self-attention] 
 { \label{self_attention}   
  
\includegraphics[width=0.48\columnwidth]{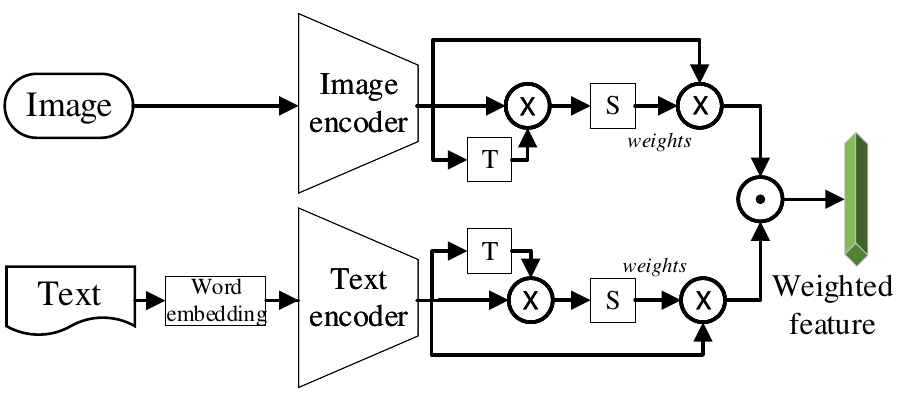}   
  }
  \vspace{-1.5em}
\caption{ Diagram for different types of attention mechanisms used in deep multimodal learning.}
\label{Attention_types}
\end {figure}

\textcolor{black}{ Attention are commonly used mechanisms to tackle this issue and have been widely incorporated into various multimodal tasks, such as visual question answering \cite{wu2018object}\cite{singh2018attention}\cite{patro2018differential}
\cite{yu2018beyond}\cite{bai2018deep}\cite{nguyen2018improved}\cite{qiao2018exploring}\cite{lu2016hierarchical}\\
\cite{chen2018factual}, image captioning \cite{chen2018show}\cite{Song2019connectingLanguage}\cite{wang2020learning}\cite{Yao2018ExploringVR}\cite{wang2018gated}\cite{gan2019multi}, and cross-modal retrieval \cite{huang2019bi}\cite{zhang2018attention}\cite{yu2020learning}. In principle, the attention mechanisms compute different weights (or importances) according to relevances between two global (or local) multimodal features and assign different importances to these features. Thereby, the networks will be more targeted at the sub-components of the source modality--regions of an image or words of a sentence. To further explore the relevances between two modalities, the attention mechanisms are adopted on multi-level feature vectors \cite{Xu2016AskAA}\cite{wang2018gated}, employed in a hierarchical scheme \cite{lu2016hierarchical}\cite{agrawal2018don}, and incorporated with graph networks for modelling semantic relationships \cite{wang2020learning}. }

To elaborate on the current ideas and trends of attention algorithms, we categorize this popular mechanism into different types. According to objective computing vectors, we categorize the current attention algorithms into four types: visual attention, textual attention, co-attention, and self-attention. Their diagrams are introduced in Figure \ref{Attention_types}. \textcolor{black}{We further categorize the attention algorithms into single-hop and multiple-hop (\emph{i.e.} stacked attention) according to the iterations of importance calculation.}

\textcolor{black}{\textbf{Visual attention}. As shown in Figure \ref{Visual_attention}, visual attention schemes are used in scenarios where text features (\emph{e.g.} from a query question) are used as context to compute their co-relevance with image features, and then the relationships are used to construct a normalized weight matrix. Subsequently, this matrix is applied to original image features to derive text-guided image features using element-wise multiplication operation (linear operation). The weighted image features have been aligned by the correlation information between image and text. Finally, these aligned multimodal features are utilized for prediction or classification.} This idea is common in multimodal feature learning \cite{chen2018show}\cite{anderson2018bottom}\cite{Song2019connectingLanguage}\cite{Yao2018ExploringVR}\cite{Xu2016AskAA}\cite{wu2018object}\cite{singh2018attention}\cite{patro2018differential}\cite{bai2018deep}\cite{chen2018factual}\cite{wang2018gated} and has been incorporated to get different text-guided features. For example, Anderson et al. \cite{anderson2018bottom}\cite{bai2018deep} employ embedded question features to highlight the most relevant image region features in visual question answering. The predicted answers are more accurately related to the question type and image content. Visual attention is widely used to learn features from two modalities.

\textbf{Textual attention}. Compared to visual attention, the textual attention approach is relatively less adopted. As shown in Figure \ref{Textual_attention}, it has an opposite computing direction \cite{qi2018cross}\cite{li2016visual}\cite{santoro2017simple}. The computed weights are based on text features to obtain relevances for different image regions or objects. According to the work \cite{zhang2019information}, the reason why textual attention is necessary is that text features from the multimodal models often lack detailed information for a given image. Meanwhile, the application of textual attention is less dominant as it is harder to capture semantic relevances between abstract text data and image data. Moreover, image data has always contained more irrelevant content for similar text. In other words, the text might describe only some parts within an image.

\textbf{Co-attention}. As shown in Figure \ref{Co_attention}, co-attention algorithm is viewed as a combination of visual attention and textual attention, which is an option to explore the inter-modality correlations \cite{huang2019bi}\cite{nguyen2018improved}\cite{lu2016hierarchical}\\ \cite{fan2018stacked}\cite{nam2017dual}\cite{gan2019multi}\cite{peng2018dynamic}\cite{li2017identity}\cite{yu2020learning}. Co-attention is a particular case of joint feature embedding in which image and text features are usually treated symmetrically. \textcolor{black}{ Co-attention in a bi-directional way is beneficial for spatial-semantic learning. As an example, Nguyen et al. \cite{nguyen2018improved} introduce a dense symmetric co-attention method to improve the fusion performance of image and text representations for VQA. In their method, features are sampled densely to fully consider each interaction between any word in question and any image region. Similarly, Huang et al. \cite{huang2019bi} also employ this idea to capture underlying fine-granularity correlations for image-text matching, \textcolor{black}{and Ding et al. \cite{ding2020stimulus} capture similar fine-granularity by two types of visual-attention for image captioning}. Meanwhile, several other works explore different formations of co-attention. For instance, Lu et al. \cite{lu2016hierarchical} explore co-attention learning in a hierarchical fashion where parallel co-attention and alternating co-attention are employed in VQA. The method aims at modelling the question hierarchically at three levels to capture information from different granularities. Integrating image feature with hierarchical text features may vary dramatically so that the complex correlations are not fully captured \cite{yu2017multi}. Therefore, Yu et al. \cite{yu2017multi}\cite{yu2018beyond} develop the co-attention mechanism into a generalized Multi-modal Factorized High-order pooling (MFH) block in an asymmetrical way. Thereby, higher-order correlations of multi-modal feature achieve more discriminative image-question representation and further result in significant improvement on the VQA performance.}

\textbf{Self-attention}. Compared to the co-attention algorithm, self-attention, which considers the intra-modality relations, is less popular in deep multimodal learning. As intra-modality relation is complementary to the inter-modality relation, its exploration is considered improving the feature learning capability of deep networks. \textcolor{black}{ For example, in the VQA task, the correct answers are not only based on their associated words/phrases but can also be inferred from related regions or objects in an image. Based on this observation, a self-attention algorithm is proposed for multimodal learning to enhance the complementary between intra-modality relations and the inter-modality relations \cite{yu2018beyond}\cite{zhang2018attention}\cite{peng2018dynamic}\cite{wang2020dynamic}. Self-attention has been used in different ways. For example, Gao et al. \cite{peng2018dynamic} combine the attentive vectors form self-attention with co-attention using element-wise product. The linear modelling method for inter- and intra-modality information flow lead to less effectiveness since the complex correlations cannot be fully learned. Therefore, more effective strategies are introduced. Yu et al. \cite{yu2018beyond} integrate text features from the self-attention block by Multimodal Factorized Bilinear (MFB) pooling approach rather than linear method to produce joint features. Differently, Zhang et al. \cite{zhang2018attention} propose a learnable combination scheme in which they employ a self-attention algorithm to extract the image and text features separately. Then these attended and unattended features are trained in an adversary manner.}

It is important to note that when these four types of attention mechanisms are applied, they can be used to highlight the relevances between different image region features and word-level, phrase-level or sentence-level text features. These different cases just need region/object proposal networks and sentence parsers. When multi-level attended features are concatenated, the final features are more beneficial for content understanding in multimodal learning. 

\textcolor{black}{ As for single-hop and multiple-hop (stacked) attention, the difference lies in whether the attention ``layer'' will be used one or more times. The four mentioned attention algorithms can be applied in a single-hop manner where the relevance weights between image and text features are computed once only. However, for multiple-hop scenarios, the attention algorithm is adopted hierarchically to perform coarse-to-fine feature learning, that is, in a stacked way \cite{yu2020learning}\cite{lu2016hierarchical}\cite{nam2017dual}\cite{Xu2016AskAA}\cite{yang2016stacked}\cite{singh2018attention}\cite{fan2018stacked} .  For example, Xu et al. \cite{Xu2016AskAA} introduce two-hop spatial attention learning for VQA. The first hop focuses on the whole and the second one focuses on individual words and produces word-level features. Yang et al. \cite{yang2016stacked} also explore multiple attention layers in VQA thereby the sharper and higher-level attention distributions will contribute refined query features for predicting more relevant answers. Singh et al. \cite{singh2018attention} achieve marginal improvements using ``attention on attention'' framework in which the attention module is stacked in parallel and for image and text feature learning. Nevertheless, a stacked architecture has tendency for gradient vanishing \cite{fan2018stacked}. Regarding this, Fan et al. \cite{fan2018stacked} propose stacked latent attention for VQA. Particularly, all spatial configuration information contained in the intermediate reasoning process is retained in a pathway of convolutional layers so that the vanishing gradient problem is tackled. }

\textcolor{black}{ In summary, to better understand the content in visual and textual modality, attention mechanisms provide a pathway for aligning the multimodal semantic correlations. With different multimodal applications, attention mechanisms (single-hop or multiple-hop) can have different benefits. To this end, we briefly make a comparison for single-hop and multiple-hop with respect to their advantages, disadvantages, and the applicable scenarios in Table \ref{single_multi_hop_attention}. 
}

\begin{table*}[ht]
\caption{\textcolor{black}{Brief comparisons of two attention categories }}\label{single_multi_hop_attention}
\vspace{-0.5em}
\centering
\scriptsize
\begin{tabular}{!{\vrule width0.1bp}p{0.5cm}|p{3.5cm}|p{4.0cm}|p{5.0cm}!{\vrule width0.1bp}}
\hline
  \multicolumn{1}{|c|}{ \textcolor{black}{\textbf{Hop(s)}} }  & \multicolumn{1}{|c|}{ \textbf{\textcolor{black}{Advantages}} } & \multicolumn{1}{|c|}{ \textbf{\textcolor{black}{Disadvantages}}}  & \multicolumn{1}{|c|}{ \textbf{\textcolor{black}{Applicable scenarios}}} \\
\hline
  \multirow{3}*{\textcolor{black}{Single}}  & \textcolor{black}{More straightforward and training effective since the visual-textual interaction occurs a single time} & \textcolor{black}{Less focused on complex relations between words. Insufficient to locate words or features on complicated sentences}	&  \textcolor{black}{No explicit constraints for visual attention. Suitable for capturing relations in short sentences as tends to be paid much to the most frequently words.}\\  
\hline
 \multirow{6}*{ $ \!\!\!\! $ \textcolor{black}{Multiple}}   & \textcolor{black}{More sophisticated and accurate, especially for complicated sentences. Each iteration provides newly relevant information to discover more fine-grained correlations between image and text.} & \textcolor{black}{Less training effective due to re-assigning attention weights multiple times. Sharing structures and parameters leads to attention bias (similar attention weights in all hops). Might suffer from the gradient vanishing problem \cite{fan2018stacked}.} & \textcolor{black}{Beneficial for multimodal learning involved long sentences. More suitable for sentence embedding in text classification or machine translation tasks. Beneficial for combining with memory networks due to the repeatedly or iteratively information extraction process.}  \\
\hline
\end{tabular}
\end{table*}

\vspace{-1em}
\subsection*{4.3 Common Latent Space Learning}
\vspace{-0.5em}

\textcolor{black}{ As illustrated in Figure \ref{The_pipeline_of_deep_multimodal_research}, feature extractors (\emph{e.g.} GCNs) would yield modality-specific presentations. In other words, these features distribute inconsistently and are not directly comparable. To this end, it is necessary to further map these monomodal features into a common latent space with the help of an embedding networks (\emph{e.g.} MLP). Therefore, common latent feature learning has been a critical procedure for exploiting multimodal correlations. In the past years, various constraint and regularization methods have been introduced into multimodal applications (see Figure \ref{Methods_Application}). In this section, we will include these ideas, such as attention mechanisms, which aim to retain similarities between monomodal image and text features.}

 According to the taxonomy methods \cite{baltruvsaitis2019multimodal}, multimodal feature learning algorithms include joint and coordinated methods. The joint feature embedding is formulated as:
 
\vspace{-2em}
\begin{equation}
\begin{aligned}
J = \mathcal{J}(x_{1},...,x_{n}, y_{1},...,y_{n})\label{joint}
\end{aligned}
\end{equation}
while coordinated feature embeddings are represented as:

\vspace{-2em}
\begin{equation}
\begin{aligned}
F = \mathcal{F}(x_{1},...,x_{n}) \thicksim \mathcal{G}(y_{1},...,y_{n}) = G\label{coordinated}
\end{aligned}
\end{equation}
where $ J $ refers to the jointly embedded features, $ F $  and $ G $ denote the coordinated features. $ x_{1},..., x_{n} $ and $ y_{1},..., y_{n} $ are $n$-dimension monomodal feature representations from two modalities (\emph{i.e.} image and text). The mapping functions $ \mathcal{J}(\cdot) $, $ \mathcal{F}(\cdot) $ and $ \mathcal{G}(\cdot) $ denote the deep networks to be learned, ``$\thicksim$'' indicates that the two monomodal features are separated but are related by some similarity constraints (\emph{e.g.} DCCA \cite{andrew2013deep} ).

\vspace{-1em}
\subsubsection*{ 4.3.1 Joint Feature Embedding}
\vspace{-0.5em}

In deep multimodal learning, joint feature embedding is a straightforward way in which monomodal features are combined into the same presentation. The fused features are used to make a classification in cross-modal retrieval \cite{wang2018joint}. It also can be used for performing sentence generation in VQA \cite{anderson2018bottom}\cite{cui2018learning}\cite{Li2018VQAEEE}.

\textcolor{black}{ In early studies, some basic methods are employed for joint feature embedding such as feature summation, feature concatenation \cite{Han2016StackGAN}\cite{HanZ2017StackGAN}\cite{Tao18attngan}, and element-wise inner product \cite{Xiong2016DynamicMN}\cite{nguyen2018improved}, the resultant features are then fed into a multi-layer perceptron to predict similarity scores. These approaches construct a common latent space for features from different modalities but cannot preserve their similarities while fully understanding the multimodal content. Alternatively, more complicated bilinear pooling methods \cite{gao2016compact} are introduced into multimodal research.  For instance, Multimodal Compact Bilinear (MCB) pooling is introduced \cite{fukui2016multimodal} to perform visual question answering and visual grounding. However, the performance of MCB is based on a higher-dimensional space. Regarding this demerit, Multimodal Low-rank Bilinear pooling \cite{kim2016hadamard}\cite{qiao2018exploring} and Multimodal Factorized Bilinear pooling \cite{yu2017multi} are proposed to overcome the high computational complexity when learning joint feature. Moreover, Hedi et al.  \cite{ben2017mutan} introduce a tensor-based
Tucker decomposition strategy, MUTAN, to efficiently parameterized bilinear interactions between visual and textual representations so that the model complexity is controlled and the model size is tractable. In general, to train an optimal model to understand semantic correlations, classification-based objective functions \cite{dash2017tac}\cite{odena2017conditional} and regression-based objective functions \cite{Han2016StackGAN}\cite{Tao18attngan} are commonly adopted.}

Bilinear pooling methods are based on outer products to explore correlations of multimodal features. \textcolor{black}{Alternatively, neural networks are used for jointly embedding features since its learnable ability for modelling the complicated interactions between image and text. For instance, auto-encoder methods, as shown in Figure \ref{Bimodal_autoencoder}, are used to project image and text features with a shared multi-layer perceptron (MLP). The similar multimodal transformer introduced in \cite{wang2020dynamic} constructs a unified joint space for image and text.} In addition, sequential networks are also adopted for the latent space construction. \textcolor{black}{Take visual question answering as an example, based on the widely-used ``encoder-decoder'' framework, image features extracted from the encoder are fed into the decoder (\emph{i.e.} RNNs \cite{zhu2018image}), and finally combined with text features to predict correct answers \cite{wu2017visual}\cite{zhao2018multi}\cite{chen2019improving}\cite{Ma2018VisualQA}. There are several ways to combine features. Image features can be viewed as the first ``word'' and concatenate all real word embeddings from the sentences.} Alternatively, image features can be concatenated with each word embedding then fed them into RNNs for likelihood estimation. Considering the gradient vanishing in RNNs, CNNs are used to explore complicated relations between features \cite{zheng2017dual}\cite{gao2018question}. For example, convolutional kernels are initialized under the guidance of text features. Then, these text-guided kernels operate on extracted image features to maintain semantic correlations \cite{gao2018question}.

\textcolor{black}{ The attention mechanisms in Section 4.2.3 can also be regarded as a kind of joint feature alignment method and are widely used for common latent space learning (see Figure \ref{Methods_Application}). Theoretically, these feature alignment schemes aim at finding relationships and correspondences between instances from visual and textual modalities \cite{baltruvsaitis2019multimodal}\cite{zhang2019information}. In particular, the mentioned co-attention mechanism is a case of joint feature embedding in which image and text features are usually treated symmetrically \cite{wang2018joint}. The attended multimodal features are beneficial for understanding the inter-modality correlations. Attention mechanisms for common latent space learning can be applied in different formations, including bi-directional \cite{nguyen2018improved}\cite{huang2019bi}, hierarchical \cite{lu2016hierarchical}\cite{yu2017multi}\cite{yu2018beyond}, and stacked \cite{lu2016hierarchical}\cite{nam2017dual}\cite{Xu2016AskAA}\cite{yang2016stacked}. More importantly, the metrics for measuring similarity are crucial in attentive importance estimation. For example, the importance estimation bu simple linear operation \cite{lu2016hierarchical} may fail to capture the complex correlations between visual and textual modality while the Multi-modal Factorized High-order pooling (MFH) method can learn higher-order semantic correlations and achieve marginal performance.  }

To sum up, joint feature embedding methods are basic and straightforward ways to allow learning interactions and perform inference over multimodal features. Thus, joint feature embedding methods are more suitable for situations where image and text raw data are available during inference, and joint feature embedding methods can be expanded into situations when more than two modalities are present. However, for content understanding among inconsistently distributed features, as reported in previous work \cite{wu2017visual}, there is potential for improvement in the embedding space.

\vspace{-1em}
\subsubsection*{4.3.2 Coordinated Feature Embedding}
\vspace{-0.5em}

\textcolor{black}{Instead of embedding features jointly into a common space, an alternative method is to embed them separately but with some constraints on features according to their similarity (\emph{i.e.} coordinated embedding).  For example, the above-noted reconstruction loss in auto-encoders can be used to constraining multimodal feature learning in the common space \cite{wu2018learning}\cite{zhan2018comprehensive}. Using traditional canonical correlation analysis \cite{hotelling1992relations}, as an alternative, the correlations between two kinds of features can be measured and then maintained \cite{tommasi2019combining}\cite{andrew2013deep}. To explore semantic correlation in a coordinated way, generally, there are two commonly used categories: classification-based methods and verification-based methods.}

\textcolor{black}{For classification-based methods when class label information is available, these projected image and text features in the common latent space are used for label prediction \cite{wang2017adversarial}\cite{wang2019learning}\cite{li2018self}\cite{peng2018dynamic}. Cross-entropy loss between the inference labels and the ground-truth labels is computed to optimize the deep networks, see Figure \ref{The_pipeline_of_deep_multimodal_research}, via the back-propagation algorithm. For classification-based methods, class labels or instance labels are needed. They map each image feature and text feature into a common space and guarantee the semantic correlations between two types of features. Classification-based methods mainly concern the image-text pair with the same class label. For the image and unmatched text (vice versa), classification-based methods have less constraints.}

\textcolor{black}{Different from classification-based methods, the commonly used verification-based methods can constrain both the matched image-text pairs (similar or have the same class labels) and unmatched pairs  (dissimilar or have the different class labels). Verification-based methods are based on metric learning among multimodal features. Given similar/dissimilar supervisory information between image and text, these projected multimodal features should be mapped based on their corresponding similar/dissimilar information. In principle, the goal of the deep networks is to make similar image-text features close to each other while mapped dissimilar image-text features further away from each other. Verification-based methods include pair-wise constraint and triplet constraint, both of which form different objective functions.}

\textcolor{black}{ For pair-wise constraint, the key point lies in constructing an inference function to infer similarity of features. For example, Cao et al. \cite{cao2016correlation} use matrix multiplication to compute the pair-wise similarity. In other examples, Cao et al. \cite{cao2018cross}\cite{li2018self}\cite{jiang2017deep} construct a Bayesian network, rather than a simple linear operation, to preserve the similarity relationship of image-text pairs.  In addition, triplet constraint is also widely used for building the common latent space. Typically, bi-directional triplet loss function is applied to learn feature relevances between two modalities \cite{wang2018joint}\cite{liu2018show}\cite{deng2018triplet}\cite{wang2019learning}\cite{zhang2018attention}\cite{qi2018cross}\cite{wang2017adversarial}. Inter-modality correlations are learned well when triplet samples interchange within image and text. However, a complete deep multimodal model should also be able to capture intra-modality similarity, which is a complementary part for inter-modality correlation. Therefore, several works consider combining intra-modal triplet loss in feature learning in which all triplet samples are from the same modality (\emph{i.e.} image or text data) \cite{wang2018joint}\cite{deng2018triplet}\cite{patro2018differential}\cite{peng2018dynamic}.}

\textcolor{black}{These classification-based and verification-based approaches are widely used for deep multimodal learning. Although the verification-based methods overcome some limits of classification-based methods, they still face some disadvantages such as the negative samples and margin selection, which inherit from metric learning \cite{zhang2018deep}. Recently, new ideas on coordinated feature embedding methods have combined adversarial learning, reinforcement learning, cycle-consistent constraints to pursue high performance. Several representative approaches are shown in Figure \ref{Methods_Application}.}

\textbf{Combined with adversarial learning}. Classification- and verification-based methods focus on the semantic relevance between similar/dissimilar pairs. Adversarial learning focuses on the overall distributions of two different modalities instead of just focusing on each pair. \textcolor{black}{The primary idea in GANs is to determine whether the input image-text pairs are matched \cite{Johnson2018Image}\cite{Hong2018Inferring}\cite{He2018An}\cite{wu2020augmented}. }

\textcolor{black}{ In new ideas of adversarial learning for multimodal learning, an implicit generator and a discriminator are designed with competitively goals (\emph{i.e.} the generator enforces similar image-text features be close while the discriminator separates them into two clusters). Therefore, the aim of adversarial learning is not to make a binary classification (``True/False''), but to train two groups of objective functions adversarially, it will enable the deep networks with powerful ability and focus on holistic features. For example, in recent works \cite{wang2017adversarial}\cite{li2018self}\cite{xu2018modal}\cite{zhang2018sch}\cite{he2017unsupervised}, a modality classifier is constructed to distinguish the visual modality and textual modality according to the input multimodal features. This classifier is trained adversarially with other sub-networks which constrain similar image-text feature to be close. Furthermore, adversarial learning is also combined with a self-attention mechanism to obtain attended regions and unattended regions. This idea is imposed on the formation of a bi-directional triplet loss to perform cross-modal retrieval \cite{zhang2018attention}.}

\textbf{Combined with reinforcement learning}. Reinforcement learning has been incorporated into deep network structures (\emph{e.g.} encoder-decoder framework) for image captioning \cite{rennie2017self}\cite{liu2018context}\cite{liu2018show}\cite{ren2017deep}\cite{zhao2018multi}\cite{gao2019self}\cite{Liu2017Improved}, visual question answering \cite{Wang2019ADR}\cite{wu2018you} and cross-modal retrieval \cite{zhang2018sch}. Because reinforcement learning avoids exposure bias \cite{liu2018show}\cite{liu2018context} and non-differentiable metric issue \cite{rennie2017self}\cite{liu2018show}. It is adopted to promote multimodal correlation modeling. To incorporate reinforcement learning, its basic components are defined (\emph{i.e.} ``agent'', ``environment'', ``action'', ``state'' and ``reward''). Usually, the deep models such as CNNs or RNNs are viewed as the ``agent'', which interacts with an external ``environment'' (\emph{i.e.} text features and image features), while the ``action'' is the prediction probabilities or words of the deep models, which influence the internal ``state'' of the deep models (\emph{i.e.} the weights and bias). The ``agent'' observes a ``reward'' to motivate the training process. The ``reward'' is an evaluation value through measuring the difference between the predictive distribution and ground-truth distribution. For example, the ``reward'' in image captioning is computed from the CIDEr (Consensus-based Image Description Evaluation) score of a generated sentence and a true descriptive sentence. The ``reward'' plays an important role for adjusting the goal of predictive distribution towards the ground-truth distribution.

Reinforcement learning is commonly used in generative models in which image patch features or word-level features are regarded as sequential inputs. When incorporating reinforcement learning into deep multimodal learning, it is important to define an algorithm to compute the expected gradients and the ``reward'' as a reasonable optimization goal. 

For the first term, the expected gradients, REINFORCE algorithm \cite{williams1992simple} is widely used as a policy gradient method to compute gradients, then to update these ``states'' via back-propagation algorithms \cite{rennie2017self}\cite{liu2018context}\cite{liu2018show} \\ \cite{zhang2018sch}\cite{gao2019self}\cite{wu2018you}\cite{Liu2017Improved}\cite{gu2018look}. \textcolor{black}{ For the second term, there are several different alternatives. For example, the difference, evaluated by the popular metric CIDEr, between the generated captions and true description sentences in image captioning is used as a ``reward'' \cite{liu2018context}\cite{gao2019self}\cite{rennie2017self}\cite{zhao2018multi}\cite{Liu2017Improved}. Instead of measuring the difference, sample similarity is more straightforward 
to track. As an example, visual-textual similarity is used as ``reward'' after deep networks are trained under the ranking loss (\emph{e.g.} a triplet loss) \cite{liu2018show}\cite{ren2017deep}\cite{zhang2018sch}. Note that, the design of triplet ranking loss function is diverse, such as in a bi-directional manner \cite{ren2017deep} or based on inter-modal triplet sampling \cite{liu2018show}. For example, Wang et al. \cite{Wang2019ADR} devise a three-stage ``reward'' for three different training scenarios according to similarity scores and binary indicators.}


\textbf{Combined with cycle-consistent constraint}. Class label information or relevance information between image and text is crucial for understanding semantic content. However, this supervisory information sometimes is not available for training deep networks. In this case, a cycle-consistent constraint is employed for unpaired image-text inputs. The basic idea of a cycle-consistent constraint is dual learning in which a closed translation loop is used to regularize the training process. This self-consistency constraint allows a predictive distribution to retain most of the correlations of the original distribution to improve the stability of network training. In principle, a cycle-consistent constraint includes a forward cycle and backward cycle. The former relies on the loss function $ F(G(X)) \approx X $, while the latter relies on another loss function $ G(F(Y)) \approx Y $. In these two functions, $ F(\cdot) $ is a mapping process from $ Y $ to $ X $ and $ G(\cdot) $ is a reversed process from $ X $ to $ Y $. Cycle-consistency has been used on several tasks such as cross-modal retrieval \cite{li2019coupled}\cite{gu2018look}\cite{wu2019cycle}\cite{qi2018crossmodal}\cite{xu2018modal}, image generation \cite{Gorti2018Text}\cite{Liu2018SemanticIS}\cite{gu2018look} and visual question answering \cite{li2018visual}.

\textcolor{black}{Cycle-consistency is an unsupervised learning method for exploring semantic correlation in the common latent space. To ensure predictive distribution and retain as many correlations as possible, the aforementioned forward and backward cycle-consistent objective functions are necessary. The feature reconstruction loss function acts as the cycle-consistency objective function. For example, Gorti et al. \cite{Gorti2018Text} utilize the cross-entropy loss between generated words and the actual words as cycle-consistency loss values to optimize the process text-to-image-to-text translation. For cross-modal retrieval tasks, Li et al. \cite{li2019coupled} adopt Euclidean distance between predictive features and reconstructed features as the cycle-consistency loss where the two cycle loss functions interact in a coupled manner to produce reliable codes.}

\textcolor{black}{Currently, the application of cycle-consistent constraints for deep multimodal learning can be categorized as structure-oriented and task-oriented. The former group focuses on making several components in a whole network into a close loop in which output of each component is used as the input for another component. Differently, task-oriented group concerns to exploit the complementary relations between tasks. Thus, there are two independent tasks (\emph{e.g.} VQA and VQG) in the close loop. }

\textcolor{black}{ For structure-oriented groups, the cycle-consistent idea is combined with some popular deep networks, such as GANs, to make some specific combinations. In these methods, image features are projected as ``text features'' and then reconstructed back to itself. Currently, the combination with GANs is a popular option since paired correspondence of modalities can be learned in the absence of a certain modality (\emph{i.e.} via generation). For example, Wu et al. \cite{wu2019cycle} plug a cycle-consistent constraint into feature projection between image and text. The inversed feature-learning process is constrained using the least absolute deviation. The whole process is just to learn a couple of generative hash functions through the cycle-consistent adversarial learning. Regarding this limit, Li et al. \cite{li2019coupled} devise an outer-cycle (for feature representation) and an inner-cycle (for hash code learning) constraint to combine GANs for cross-modal retrieval. Thereby, the objects for which the cycle-consistency loss constrains have increased. Moreover, in their method, the discriminator should distinguish if the input feature is original (viewed as \textit{True}) or generated (viewed as \textit{False}).}

\textcolor{black}{ For task-oriented groups, cycle-consistency is adopted into dual tasks. In cycle-consistency, we use an inverse process (\textit{task A} to \textit{task B} to \textit{task A}) to improve the results. When a whole network performs both tasks well, it indicates that the learned features between the tasks have captured the semantic correlations of two modalities. For example, Li et al. \cite{li2018visual} combine visual question answering (VQA) and visual question generation (VQG), in which the predicted answer is more accurate through combining image content to predict the question. In the end, the complementary relations between questions and answers lead to performance gains. For text-image translation, a captioning network is used to produce a caption which corresponds to a generated image from a sentence using GANs \cite{Gorti2018Text}. The distances between the ground truth sentences and the generated captions are exploited to improve the network further. The inverse translation is beneficial for understanding text context and the synthesized images. To sum up, there are still some questions to be explored in task-oriented ideas, such as the model parameter sharing scheme, and these implicit problems make the model more difficult to train and might encounter gradient vanishing problems, the task-oriented cycle-consistent constraint is applied to unify multi-task applications into a whole framework and attracts more research attention.}

\vspace{-1.5em}
\section*{5 Conclusion and Future Directions}
\vspace{-1em}

In this survey, we have conducted a review of recent ideas and trends in deep multimodal learning (image and text) including popular structures and algorithms. We analyzed two major challenges in deep multimodal learning for which these popular structures and algorithms target. Specifically, popular structures including auto-encoders, generative adversarial nets and their variants perform uni-directional and bi-directional multimodal tasks. Based on these popular structures, we introduced current ideas about multimodal feature extraction and common latent feature learning which plays crucial roles for better content understanding within a visual and textual modality. For multimodal feature extraction, we introduced graph convolutional networks and memory-augmented networks. For common latent feature learning, we presented the joint and coordinated feature embedding methods including the recently proposed objective functions.

\begin {figure}[!t]
\centering
 { 
   \includegraphics[width= 0.9 \columnwidth]{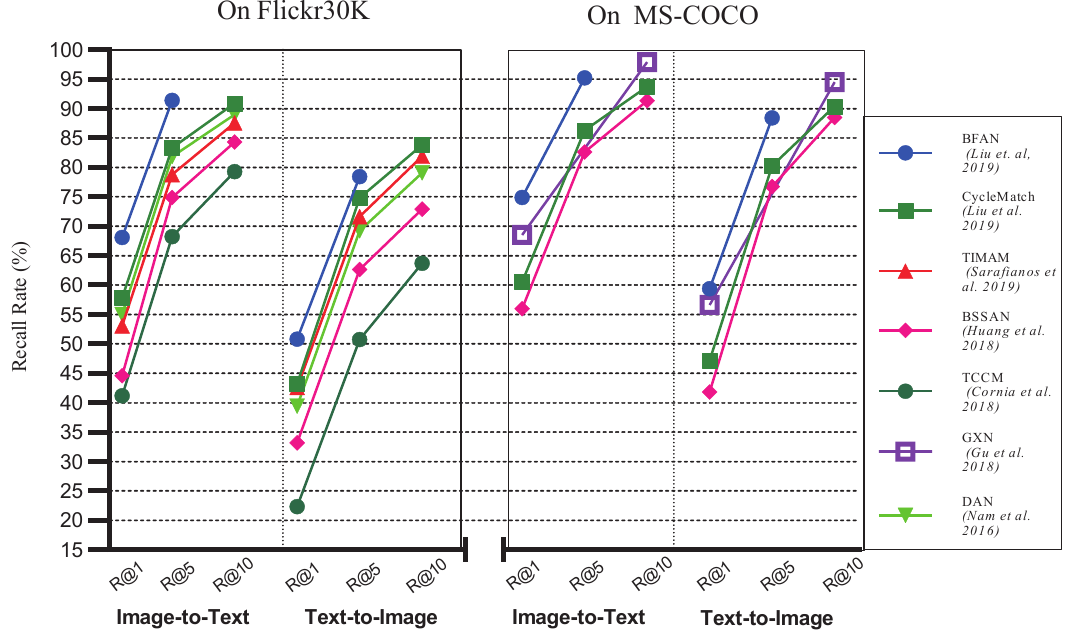} 
 }
 \vspace{-1em}
\caption{ The achieved progress of cross-modal retrieval on the Flickr30K \cite{plummer2015flickr30k} and the MS-COCO \cite{lin2014microsoft} datasets. }
\label{Flickr30k_MSCOCO}
\end {figure}

\begin {figure}[!t]
\centering
 { 
   \includegraphics[width=0.9 \columnwidth]{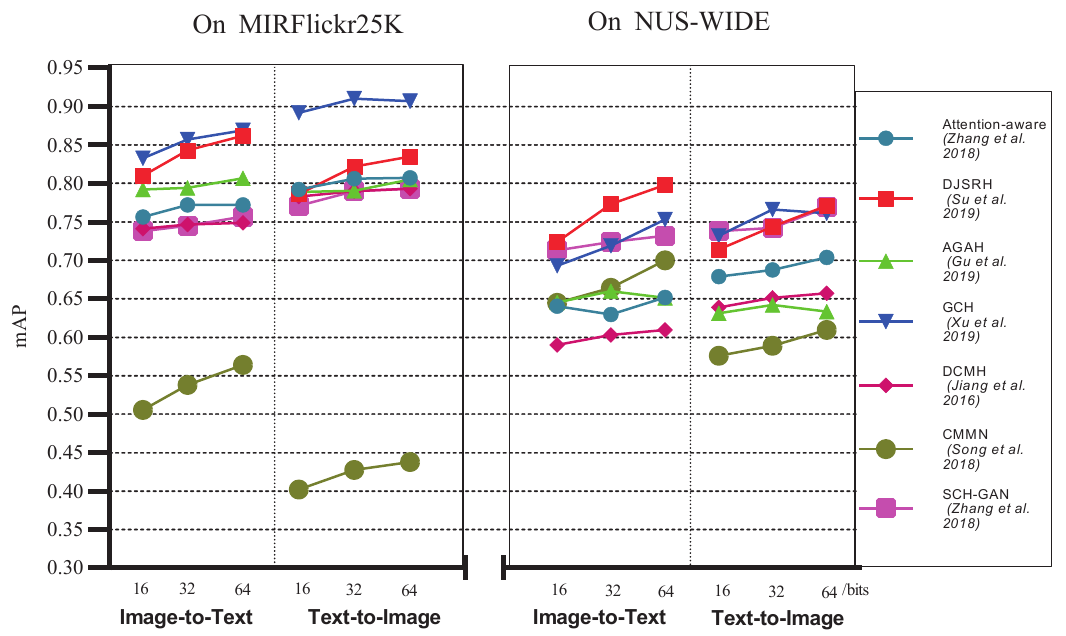} 
 }
 \vspace{-1em}
\caption{ The achieved progress of cross-modal hash retrieval on the MIRFlickr25k \cite{huiskes2008mir} and the NUS-WIDE \cite{chua2009nus} datasets. Hashing methods have higher retrieval efficiency using the binary hash codes. }
\label{Flickr25k_NUSWIDE_Hashing}
\end {figure}

The aforementioned ideas have made some progress on various multimodal tasks. For example, for cross-modal retrieval, we presented the achieved progress and state-of-the-art of recent methods on the Flickr30K \cite{plummer2015flickr30k} and the MS-COCO \cite{lin2014microsoft} datasets in Figure \ref{Flickr30k_MSCOCO}. For hashing retrieval methods, we presented the achievement on the MIRFlickr25k \cite{huiskes2008mir} and the NUS-WIDE \cite{chua2009nus} datasets in Figure \ref{Flickr25k_NUSWIDE_Hashing}. Meanwhile, other tasks, (\emph{i.e.} image generation, image captioning and VQA) are listed in Tables in the Appendix. \textcolor{black}{ As we can see from these statistics, the progress is notable in recall rate (\emph{i.e.} the fraction of queries for which the top K nearest neighbors are  retrieved correctly) and mAP (\emph{i.e.} the mean of the average precision scores for each query) in cross-modal retrieval. However, there is still room for improvement in the current limitations of multimodal content understanding.}

\textcolor{black}{ For future directions, multi-task integrated networks might be helpful and complementary for content understanding as different applications capture semantic correlations from different perspectives. Effort has been made on integrating image captioning and cross-modal retrieval tasks, image captioning and visual question answering, image generation and image retrieval. Nevertheless, these combined applications are only based on two modalities. Considering the complementary characteristic among modalities (conveying the same concept), it might be promising to fuse more than two modalities to enable machines to understand their semantic correlations. Undoubtedly, it will be more challenging for aligning these diverse data. There are some explorations in this direction. Aytar et al. \cite{aytar2017see} present a deep cross-modal convolutional network to learn a representation that is aligned across three modalities: sound, image, and text. The network is only trained with ``image + text'' and ``image + sound'' pairs. He et al. \cite{he2019new} construct a new benchmark for cross-media retrieval in which image, text, video, and audio are included. It is the first benchmark with 4 media types for fine-grained cross-media retrieval. However, this direction is still far from satisfactory. }

\textcolor{black}{Deep neural networks, including convolutional neural networks and recurrent neural networks, have made the monomodal feature extraction and multimodal feature learning end-to-end trainable. The representations from multimodal data can be automatically learned effectively, without the need of requiring expert knowledge in a certain field, which makes the process of understanding of multimodal content more intelligent. However, the disadvantages of deep networks for multimodal learning are obvious. It is well-known that the deep networks depend on a massive of multiple-modality data to train, but the less biased datasets are not so common. More importantly, deep networks for multimodal learning lacks of interpretability to some extent. Although joint embedding or coordinated embeddings methods can be utilized, it still needs to figure out which modality (or its features) plays more important role for the final content understanding. }

From a technical viewpoint, graph-based networks are an important direction for future research. Currently, graph representation is constructed within intra-modality to present the semantic relations, which can be further explored in the future. Meanwhile, the exploration of graph-based networks can be deepened by examining scalability and heterogeneity \cite{Wu2019ACS}. Finally, generation-based tasks such as image generation and image captioning are effective for unsupervised learning, since numerous labeled training data can be generated from the deep networks. Combined with reinforcement learning, the image generation process is more controllable. For example, some fine-grained attributes including texture, shape and color can be specified during deep network training. Once it understands the content between modalities, the deep network, like an agent, will synthesize photo-realistic images, which can be used in other applications. 

\vspace{-1.5em}
\section*{Acknowledgments}

This work was supported by LIACS MediaLab at Leiden University and China Scholarship Council (CSC No. 201703170183). We appreciate the helpful editing work from Dr. Erwin Bakker.

\section*{Appendix}

The progress of other multimodal application in recent years including image captioning, image generation and visual question answering. We list the topic-relevant representative works in several datasets. The evaluation metric are these common-used in each task. These approaches make sense for improving performance.

\newgeometry{left = 0.5 cm, top=3 cm}
\begin{landscape}
\begin{table}[bp]
\centering
\setlength{\tabcolsep}{1mm}
\caption{Performance of image captioning on the MS-COCO dataset \cite{lin2014microsoft}}
\begin{tabular}{p{89 pt}|c|c|c|c|c|c|c|c|c|c|c|c|c|c|c|l}
\cline{1-17}

\multicolumn{1}{c|}{\multirow{2}{*}{\textbf{Methods}}} & \multirow{2}{*}{\textbf{Year}} & \multicolumn{2}{c|}{\textbf{CIDEr}}                & \multicolumn{2}{c|}{\textbf{ROUGE-L}}              & \multicolumn{2}{c|}{\textbf{METEOR}}               & \multicolumn{2}{c|}{\textbf{BLEU1}}                & \multicolumn{2}{c|}{\textbf{BLEU2}}                & \multicolumn{2}{c|}{\textbf{BLEU3}}                & \multicolumn{2}{c|}{\textbf{BLEU4}}                & \multicolumn{1}{c}{\multirow{2}{*}{\textbf{KeyNotes}}} \\ \cline{3-16}
\multicolumn{1}{c|}{}                                  &                                & \multicolumn{1}{c|}{c5} & \multicolumn{1}{c|}{c40} & \multicolumn{1}{c|}{c5} & \multicolumn{1}{c|}{c40} & \multicolumn{1}{c|}{c5} & \multicolumn{1}{c|}{c40} & \multicolumn{1}{c|}{c5} & \multicolumn{1}{c|}{c40} & \multicolumn{1}{c|}{c5} & \multicolumn{1}{c|}{c40} & \multicolumn{1}{c|}{c5} & \multicolumn{1}{c|}{c40} & \multicolumn{1}{c|}{c5} & \multicolumn{1}{c|}{c40} & \multicolumn{1}{c}{}                                    \\  \cline{1-17}

StructCap \cite{Chen2017StructCapSS}  & 2016 & 94.3 & 95.8 & 53.5 & 68.2 & 25.0 & 33.5 & 73.1 & 90.0 & 56.5 & 81.5 & 42.4 & 70.9 & 31.6 & 59.9 & Attention + VP-tree for visual features  \\ 

 Semantic-Attn \cite{you2016image}  & 2016 & 95.3 & 94.8 & 53.4 & 68.4 & 25.1 & 34.0 & 72.4 & 90.7 & 55.8 & 82.2 & 42.3 & 71.7 & 32.0 & 60.7 & Visual attention \\ 
 
CGAN \cite{dai2017towards}  & 2017 & 102.0 & - & 52.7 & - & 24.8 & - & - & - & - & - & 39.3 & - & 29.9  & - & CGAN + Reinforcement learning  \\ 

Adaptive-Attn \cite{lu2017knowingtolook}  & 2017 & 104.2 & 105.9 & 55.0 & 70.5 & 26.4 & 35.9 & 74.8 & 92.0 & 58.4 & 84.5 & 44.4 & 74.4 & 33.6  & 63.7 & Adaptive visual attention  \\

SCST \cite{rennie2017self}  & 2017 & 114.7 & 116.7 & 56.3 & 70.7 & 27.0 & 35.5 & 78.1 & 93.7 & 61.9 & 86.0 & 47.0 & 75.9 & 35.2  & 64.5 &  Reinforcement learning  \\ 

RL-GAN \cite{yan2018image}  & 2018 & - & - & - & - & 24.3 & - & 71.6 & - & 51.8 & - & 37.1 & - & 26.5 & - &  GAN + Reinforcement learning   \\

SOT \cite{chen2018show}  & 2018 & 106.1 & 108.7 & 55.5 & 69.9 & 25.9 & 34.2 & 78.7 & 93.5 & 61.5 & 85.5 & 46.5 & 74.8 & 34.5 & 63.3 &  Visual attention   \\ 

SR-PL \cite{liu2018show}  & 2018 & 117.1 & - & 57.0 & - & 27.4 & - & 80.1 & - & 63.1 & - & 48.0 & - & 35.8 & - &  Reinforcement learning   \\

Up-Down \cite{rennie2017self}  & 2018 & 117.9 & 120.5 & 57.1 & 72.4 & 27.6 & 36.7 & 80.2 & 95.2 & 64.1 & 88.8 & 49.1 & 79.4 & 36.9  & 68.5 &   Top-down and bottom-up attention  \\

CAVP \cite{liu2018context}  & 2018 & 121.6 & 123.8 & 58.2 & 73.1 & 28.1 & 37.0 & 80.1 & 94.9 & 64.7 & 88.8 & 50.0 & 79.7 & 37.9 & 69.0 &  Reinforcement learning   \\ 

RFNet \cite{jiang2018recurrent}  & 2018 & 122.9 & 125.1 & 58.2 & 73.1 & 28.2 & 37.2 & 80.4 & 95.0 & 64.9 & 89.3 & 50.1 & 80.1 & 38.0 & 69.2 &  Visual attention   \\ 

iVQA \cite{liu2018inverse}  & 2018 & 168.2 & - & 46.6 & - & 20.1 & - & 42.1 & - & 32.0 & - & 25.3 & - & 20.5 & - &  Reinforcement learning   \\ 

UnsupervisedIC \cite{feng2019unsupervised}  & 2019 & 54.9 & - & 43.1 & - & 17.9 & - & 58.9 & - & 40.3 & - & 27.0 & - & 19.6 & - &  VAE + GAN (unsupervised)   \\ 

Graph-align \cite{gu2019unpaired}  & 2019 & 69.5 & - & - & - & 20.9 & - & 67.1 & - & 47.8 & - & 32.3 & - & 21.5 & - &  VAE +  Graph embed + cycle (unpaired) \\

Self-critical \cite{gao2019self}  & 2019 & 112.6 & 115.3 & 56.1 & 70.4 & 26.9 & 35.4 & 77.6 & 93.1 & 61.3 & 86.1 & 46.5 & 76.0 & 34.8 & 64.6 &  Reinforcement learning \\ 

PAGNet \cite{Song2019connectingLanguage}  & 2019 & 118.6 & - & 58.6 & - & 30.4 & - & 83.2 & - & 62.8 & - & 46.3 & - & 40.8 & - & Attention + Reinforcement learning \\

RL-CGAN \cite{chen2019improving}  & 2019 & 123.1 & 124.3 & 59.0 & 74.4 & 28.7 & 38.2 & 81.9 & 95.6 & 66.3 & 90.1 & 51.7 & 81.7 & 39.6 & 71.5 &  GAN + Reinforcement learning   \\ 

SGAE \cite{Yang2018AutoEncodingSG}  & 2019 & 123.8 & 126.5 & 58.6 & 73.6 & 28.2 & 37.2 & 81.0 & 95.3 & 65.6 & 89.5 & 50.7 & 80.4 & 38.5 & 69.7 &  VAE + graph embedding   \\

\hline 
           
\end{tabular}
\end{table}


\begin{table}[h]

\vspace{-1.0em}
\setlength{\tabcolsep}{1mm}
\caption{Performance of image generation}

\begin{tabular}{c|c|c|c|c|c|c|c|l}
\cline{1-9}
\multicolumn{1}{c|}{\multirow{2}{*}{\textbf{Methods}}}     & \multirow{2}{*}{\textbf{Year}} & \multicolumn{2}{c|}{\textbf{Caltech-UCSD Birds200}} & \multicolumn{2}{c|}{\textbf{Oxford Flowers102}} & \multicolumn{2}{c|}{\textbf{MS-COCO}} & \multicolumn{1}{c}{\multirow{2}{*}{\textbf{KeyNotes}}} \\ \cline{3-8}
  &   & IS  & FID   & IS  & FID  & IS   & FID  &    \\ \cline{1-9}
                     
GAN-INT-CLS \cite{reed2016generative}  & 2016   & 2.88$ \pm $.04  & -   & 2.66$ \pm $.03  & - & 7.88$ \pm $0.07 & - & Vanilla GAN for image generation  \\

TAC-GAN \cite{dash2017tac}  & 2017   & -   & -   & 3.45$ \pm $.05  & -   & - & -   & Discriminator learns class information  \\

GAWWN \cite{articleReed}   & 2017 & 3.60$ \pm $.07    & -  & -   & -   & -  & -   & Conditional objection location is learned    \\

StackGAN \cite{Han2016StackGAN}    & 2017   & 3.70$ \pm $.04  & 51.89   & 3.20$ \pm $.01 & 55.28  & 8.45$ \pm $.03  & 74.05  & GAN in a stacked structure   \\

StackGAN++ \cite{HanZ2017StackGAN}   & 2018    & 4.04$ \pm $.05  & 15.3   & 3.26$ \pm $.01    & 48.68  & 8.30$ \pm $.10   & 81.59  & GAN in a tree-like structure  \\

HDGAN \cite{Zhang2018Photographic}   & 2018   & 4.15$ \pm $.05   & -  & 3.45$ \pm $.07  & -  & 11.86$ \pm $0.18  & -  & GAN in a hierarchically-nested structure   \\

AttnGAN \cite{Tao18attngan}   & 2018  & 4.36$ \pm $.03   & -   & -  & -  & 25.89$ \pm $.47 & - & Attentional generative network  \\

Scene graphs \cite{Johnson2018Image}   & 2019  & -   & -  & -    & -   & 7.3$ \pm $0.1 & - &  Graph convolution for graphs from text \\

Obj-GANs \cite{li2019object}    & 2019  & -  & -    & -   & -  & 27.37$ \pm $0.22  & 25.85   & Attentive generator and object-wise discriminator \\

vmCAN \cite{zhang2018text}  & 2019   & -  & -   & -   & -   & 10.36$ \pm $0.17  & -  & Visual-memory method in GAN  \\

AAAE \cite{xu2019adversarially}    & 2019   & -     & -   & -   & 103.46  & -   & -   & Auto-encoders + GAN for adversarial approximation \\

Text-SeGAN \cite{cha2019adversarial} & 2019  & -  & - & 4.03$ \pm $0.07   & -  & -   & -    & Semantic relevance matching in GAN   \\

DAI \cite{lao2019dual}   & 2019  & 3.58$ \pm $0.05  & 18.41$ \pm $1.07   & 2.90$ \pm $0.03  & 37.94$ \pm $0.39  & 8.94$ \pm $0.2  & 27.07$ \pm $2.55 & Dual inference mechanism disentangled variables    \\

C4Synth \cite{joseph2019c4synth}  & 2019  & 4.07$ \pm $0.13   & -  & 3.52$ \pm $0.15 & -   & -  & -  & Image generation using multiple captions \\

PPAN \cite{Gao2019Perceptual} & 2019   & 4.35$ \pm $.05  & - & 3.53$ \pm $.02  & -   & -  & -  & GAN in perceptual pyramid structure \\

MirrorGAN \cite{qiao2019mirrorgan}  & 2019    & 4.56$ \pm $0.05    & -  & -   & -  & 26.47$ \pm $0.41 & -  & Task-oriented cycle consistency + attention \\

ControlGAN \cite{li2019controllable}  & 2019  & 4.58$ \pm $0.09  & -  & -  & -   & 24.06$ \pm $0.6  & -   & Attention + region-wise attribute generation   \\

SD-GAN \cite{yin2019semantics} & 2019 & 4.67$ \pm $0.09 & - & -  & -   & 35.69$ \pm $0.5  & -  &  Disentangling high-/low-level semantics in GAN  \\

\hline   

\end{tabular}
\\
\vspace{0.5em}
\footnotesize{ \ddag \ To evaluate the identification and diverse of generated image, Inception Score (IS) and Fr\'{e}chett Inception Distance (FID) \\ are commonly used. For Inception Score, higher is better. For Fr\'{e}chet Inception Distance, lower is better. }
\end{table}

\begin{table}[!t]

\vspace{-1.0em}
\setlength{\tabcolsep}{1mm}

\caption{Performance of visual question answering on VQA 1.0 dataset \cite{agrawal2017vqa}}
\begin{tabular}{c|c|c|c|c|c|l}
\cline{1-7}
\multirow{2}{*}{\textbf{Methods}} & \multirow{2}{*}{\textbf{Year}} & \multirow{2}{*}{\textbf{\begin{tabular}[c]{@{}c@{}}Open-ended\\ test-std\end{tabular}}} & \multirow{2}{*}{\textbf{\begin{tabular}[c]{@{}c@{}}Open-ended\\ test-dev\end{tabular}}} & \multirow{2}{*}{\textbf{\begin{tabular}[c]{@{}c@{}}MC\\ test-std\end{tabular}}} & \multirow{2}{*}{\textbf{\begin{tabular}[c]{@{}c@{}}MC\\ test-dev\end{tabular}}} & \multirow{2}{*}{\textbf{KeyNotes}} \\
   &    &   &   &  &   &  \\ \cline{1-7}
 
 Smem-VQA \cite{Xu2016AskAA} & 2016  & 58.24  & 57.99   & -  & -   & Spatial memory network stores image region features \\
  
 DMN+ \cite{Xiong2016DynamicMN}    & 2016   & 60.4  & 60.3  & -  & -   & Improved dynamic memory network for VQA \\
 
 MLB \cite{kim2016hadamard}  & 2016   & 65.07    & 64.89  & 68.89 & -   & Low-rank bilinear pooling for similarity learning  \\

  MCB \cite{fukui2016multimodal} & 2016  & 66.5  & 66.7  & 70.1  & 70.2  & Multimodal compact bilinear pooling for similarity learning  \\

High-order Attn \cite{schwartz2017high}  & 2017  & -  & - & 69.3 & 69.4  & Attention mechanisms learn high-order feature correlations  \\

DAN \cite{nam2017dual}   & 2017  & 64.2   & 64.3 & 69    & 69.1   & Co-attention networks for multimodal feature learning\\

MLAN \cite{yu2017multilevel}  & 2017  & 65.3  & 65.2   & 70   & 70    & Multi-level co-attention for feature alignment \\

SVA \cite{zhu2017structuredVQA}  & 2017  & 66.1  & 66  & -  & -  & Visual attention on grid-structured image region feature learning\\

MFB \cite{yu2017multi} & 2017  & 66.6  & 66.9  & 71.4  & 71.3  & Multi-modal factorized bilinear pooling for similarity learning \\

MUTAN \cite{ben2017mutan} & 2017   & 67.36 & 67.42  & -  & -  & Multimodal tucker fusion for similarity learning \\

Graph VQA \cite{teney2017graph}  & 2017 & 70.42  & -  & 74.37   & -  & Graph representation for scene and question feature learning \\

MAN-VQA \cite{Ma2018VisualQA} & 2018  & 64.1 & 63.8   & 69.4   & 69.5  & Memory-augmented network for feature learning and matching \\ 

QGHC \cite{gao2018question} & 2018 & 65.9  & 65.89  & -  & - & Question-guided convolution for visual-textual correlations learning \\

Dual-MFA \cite{lu2018co} & 2018  & 66.09  & 66.01  & 69.97   & 70.04   & Co-attention for visual-textual feature learning \\

VKMN \cite{ Su2018LearningVK} & 2018  & 66.1  & 66 & 69.1  & 69.1   & Visual knowledge memory network for feature learning \\

CVA \cite{song2018pixels}  & 2018  & 66.2  & 65.92  & 70.41  & 70.3   & Cubic visual attention for object-region feature learning  \\

DCN \cite{nguyen2018improved}   & 2018  & 67.02  & 66.89    & -   & -   & Dense co-attention for feature fusion  \\

DRAU \cite{osman2018dual}  & 2018  & 67.16  & 66.86  & -   & -   & Recurrent co-attention for feature learning  \\

ODA \cite{wu2018object}  & 2018  & 67.97  & 67.83  & 72.23  & 72.28   & Object-difference visual attention to fuse features \\

ALARR \cite{liu2018adversarial}  & 2018  & 68.43  & 68.61   & 71.28  & 68.43 & Adversarial learning for pair-wise feature discrimination \\

DF \cite{wu2019differential}  & 2019  & 68.48  & 68.62   & 73.05  & 73.31 & Differential network for visual-question feature learning \\

Relational Encoding \cite{liu2019language}  & 2019  & 69.3  & 69.1   & -  & - & Textual attention for question feature encoding\\

DCAF \cite{liu2019densely}  & 2019  & 70.0  & 69.9   & -  & - & Dense co-attention for feature fusion \\

\hline
     
\end{tabular}
\end{table}
\end{landscape}

\newgeometry{left = 3 cm, top=3 cm}
\bibliography{mybibfile}

\end{document}